\documentclass[nonblindrev]{informs3}

\OneAndAHalfSpacedXI

\usepackage{natbib}
 \bibpunct[, ]{(}{)}{,}{a}{}{,}%
 \def\bibfont{\small}%

 \usepackage{algorithm}
\usepackage{algpseudocode}
\usepackage[dvipsnames]{xcolor}
\usepackage{scrextend}
\usepackage[english]{babel}
\usepackage{textcmds}

\TheoremsNumberedThrough     
\ECRepeatTheorems

\EquationsNumberedThrough    

\MANUSCRIPTNO{(HBS Working Paper 21-016, First Draft: August 2020)}

\begin{document}


 \RUNAUTHOR{Valavi et al.} 

\RUNTITLE{Time and the Value of Data}

\TITLE{Time and the Value of Data}

\ARTICLEAUTHORS{%
\AUTHOR{Ehsan Valavi}
\AFF{Harvard Business School, Boston, MA 02163, \EMAIL{evalavi@hbs.edu}} 
\AUTHOR{Joel Hestness}
\AFF{Cerebras Systems, Sunnyvale, CA 94085,  \EMAIL{joel@cerebras.net}}

\AUTHOR{Newsha Ardalani}
\AFF{Baidu Research, Sunnyvale, CA 94085, \EMAIL{newsha@baidu.com}}
\AUTHOR{Marco Iansiti}
\AFF{Harvard Business School, Boston, MA 02163, \EMAIL{miansiti@hbs.edu}}} 
\ABSTRACT{%
Managers often believe that collecting more data will continually improve the accuracy of their machine learning models. However, we argue in this paper that when data lose relevance over time, it may be optimal to collect a limited amount of recent data instead of keeping around an infinite supply of older (less relevant) data. In addition, we argue that increasing the stock of data by including older datasets may, in fact, damage the model's accuracy. Expectedly, the model's accuracy improves by increasing the flow of data (defined as data collection rate); however, it requires other tradeoffs in terms of refreshing or retraining machine learning models more frequently.

Using these results, we investigate how the business value created by machine learning models scales with data and when the stock of data establishes a sustainable competitive advantage. We argue that data's time-dependency weakens the barrier to entry that the stock of data creates. As a result, a competing firm equipped with a limited (yet sufficient) amount of recent data can develop more accurate models. This result, coupled with the fact that older datasets may deteriorate models' accuracy, suggests that created business value doesn't scale with the stock of available data unless the firm offloads less relevant data from its data repository. Consequently, a firm's growth policy should incorporate a balance between the stock of historical data and the flow of new data.

We complement our theoretical results with an experiment. In the experiment, we use the simple yet widely used machine learning task known as next work prediction. We empirically measure the loss in the accuracy of a next word prediction model trained on datasets from various time periods. Our empirical measurements confirm the economic significance of the value decline over time. For example, 100MB of text data, after seven years, becomes as valuable as 50MB of current data for the next word prediction task. 

}%



\maketitle

%


\section{Introduction}

We witness a dramatic acceleration of digitization in firms' infrastructure, products, and services. Artificial Intelligence (AI) enabled solutions are on the rise, and more than ever, data appears to be a critical strategic asset [\cite{c1},\cite{c4},\cite{c11},\cite{c17}]. As a result, in almost all industries and economic sectors, firms amass substantial volumes of user data to improve their current and future services, anticipating that it also gives them an advantage over their competitors. In regulatory debates, this accumulation of data by firms is considered to be a critical source of competitive advantage that could lead to a concentration in digital markets [\cite{c11},\cite{c18},\cite{c22},\cite{c42}]. In addition, from users perspective, there are privacy concerns [\cite{c9}] on when and how firms use the accumulated data and if in any way it can adversely harm users. Because of this aggressive data accumulation by firms, it is crucial to understand how and when the value created by AI-enabled services scales with the size of available data. Particularly, since the data accumulation process often happens over time, it is of great interest to research how the created business value changes over time, especially when the dataset is sampled from a dynamically changing environment.

Current literature on how the increase in the stock of available data scales the business value has mixed results. Managers often believe that collecting more data continually improves the accuracy of machine learning models. This belief is engrained by the statistical theories on how the accuracy of machine learning models scales with the dataset size [\cite{c12}, \cite{c25}] and the managerial theories on how the economic value created by a firm scales with the resources [\cite{c1},\cite{c4},\cite{c11}]. All these theories attest that more data is always better, and securing a vast amount of such resource leads to the firm's success in the long run. In addition, recent research hypothesizes a feedback loop [\cite{c24},\cite{c28}] between the size of available data and the quality of AI-based solutions.  \cite{c24} theorize and compare this data externality to network effects, where the value of a service or product increases in user-base size. In this \qq{data network effect} [\cite{c24},\cite{c27},\cite{c36}], more data leads to a higher accuracy of algorithms, which means better services [\cite{c25}]. Better service then leads to a higher user engagement or a larger user-base, which creates even more data. Despite these theories, empirical research [\cite{c5},\cite{c13}] finds limited or no economic significance in accumulating large datasets. For example, \cite{c13} investigates the effect of historical search data on search results' quality. They found little empirical evidence on the effectiveness of old data in the quality of search engine results. \cite{c5} also raise a similar question on the data's economies of scale for specific problems. They suggest a diminishing return to scale value model for data and argue that increasing data volume in advertisement applications does not improve the service quality. We believe that diminishing return to scale theory doesn't explain the aggressive data accumulation by firms already equipped with massive datasets.

This paper investigates how the business value changes over time when the dataset is sampled from a dynamically changing environment and how this change explains the mixed results seen in the literature. A dataset sampled from a dynamically changing environment loses relevance over time, making the created business value time-dependent. This time-dependency is referred to as concept drift in the machine learning literature. Concept drift is known to cause a deterioration in the algorithm's performance. It manifests itself as a decrease in the algorithm's accuracy score or an increase in its loss value or error. However, the extent to which it affects accuracy or loss values and how time-dependency affects a firm's data strategy is still unknown. 

Our approach to the problem has similarities and differences with machine learning and AI literature. Similar to machine learning literature, we model the change in environment with a shift in data-generating probability distributions. In contrast with the literature, we fix a model/task and vary the data generating distribution to study the effect of time-dependency on business value. In machine learning research, given a dataset, researchers alter the models to improve the accuracy score or the loss value. In addition to this difference, we also define new metrics such as \qq{equivalent size} and \qq{effectiveness} to compare the value of datasets sampled from different times. Unlike machine learning literature which reports the effect of time-dependency in accuracy scores or loss values, our measures report the effect in dataset sizes. In doing so, for any machine learning task, we first define an oracle dataset. Subsequently, we train our model on the given dataset (referring to as baseline dataset) and then measure the model's accuracy score by testing it on the oracle dataset. We then ask what size of the oracle dataset leads to a similar accuracy score if used for training the model. We call it the equivalent oracle size. Thus, we can quickly compare various datasets by comparing their equivalent sizes. Another benefit of measuring the value this way is that we can borrow terminology from economics and management research, making our findings more relatable to a broader audience. For example, if a dataset's equivalent oracle size declines over time, we call the dataset perishable.

In our theoretical setting for this paper, we investigate the effect of time-dependency for the task of learning the probability distribution and use the maximum likelihood estimation method to accomplish this task. Learning the data-generating probability distribution is a fundamental problem in the statistical learning theory. It is because we can evaluate any statistics (like expectation or variance of any quantity of interest) from the distribution. Hence, we believe that theorems and propositions we prove for this task can be, with slight modification, used for a wide variety of other tasks. We use the Maximum Likelihood Estimation (MLE) for our analysis since consistency and efficiency are essential for our mission in this paper. Consistency is critical since we model dynamically changing environments using probability distributions. Hence, for any given time, it is crucial to learn the distribution consistently. Efficiency is essential to achieving the lowest estimation variance with the smallest dataset size. Intuitively, efficiency makes MLE the most scalable method (in gaining a better accuracy score) for a fixed dataset size and hence, a wise choice by firms.

We derive several managerial and economic intuitions by comparing the value of datasets sampled from different times. We argue that due to shifts in the data generating distribution, it may be optimal for a firm to collect a more limited amount of recent data instead of keeping around an infinite supply of older data. This is a direct result of our first proposition in this paper. This proposition shows that even a perfect model trained on an infinite supply of time-dependent data may have lower accuracy than the same model trained on a recent (perfectly relevant) dataset of limited size. In other words, a less relevant dataset of infinite size has a finite (bounded) equivalent oracle size (defined as the perfectly relevant dataset in this case). Hence, a competing firm with an oracle dataset of sufficient size can easily attain a better accuracy score or loss value. This proposition has several other economic and managerial implications that we discuss later in this paper. In pursuit of comparing the value of datasets from different times, we define a substitution function that measures how much oracle size a firm gains/loses if it substitutes its baseline dataset with another dataset of the same size from a different time. We prove that substitution gain is a function of the baseline dataset size as well as time. It becomes sharper with the increase in the size of the baseline dataset, meaning that the gain/loss percentage increases as the baseline dataset size increases. As we discuss later in this paper, it has immediate implications for firms regarding training frequency, i.e., how often firms should retrain their models.

We use the machinery we developed for comparing values of datasets sampled from different times to compare values of datasets curated over a period of time. We do so by defining the \qq{equivalent time} in Proposition 3. This proposition states that, for any baseline dataset curated over a period of time, there exists an \qq{equivalent time} such that, fixing the size, a model trained on a dataset from the equivalent time produces a similar accuracy score as the model trained on the baseline dataset. As a result, we can compare datasets curated over various time periods by first calculating their equivalent times and then by comparing the oracle sizes each equivalent time produces. A direct result of this method is the introduction of offloading algorithm. Offloading algorithm removes less relevant data hoping that gain in relevance counterbalances the loss in size. We then use the offloading algorithm to argue that increasing the stock of data by including older datasets may, in fact, damage the model's accuracy, putting a firm in a disadvantageous position. Together with the increase in sharpness of substitution gain as a function of the flow of data (Number of data points in a given time or the rate of acquiring new data points), these results build the case for defining the optimal scaling and growth path for a firm. When the firm is small, the optimal growth path focuses on the stock of available data curated over time. As the flow of data increases (A firm acquires more users or user-engagement increases, for example), the firm offloads older data and focuses on the flow of data as the primary value driver.

To confirm the economic significance of our findings, we empirically measure the decline in the value of data for the next word prediction task. It is a widely used machine learning task with applications in auto-completion software in cellphones and the search recommendations in search engines. In our experiment, we use a user-generated text dataset from Reddit.com [\cite{c19}]. We divide this dataset into smaller datasets based on data points' sampling time (Month-year format). We then train a variation of GPT-2 from OpenAI [\cite{c23}] on each of these smaller datasets and measure their equivalent sizes over time. Our measurements confirm the economic significance of time-dependency as we show that in roughly seven years, 100MB of text data becomes as valuable as 50MB of current data for the next word prediction task. 

Our findings can explain the mixed the result in the literature. We acknowledge that increasing the dataset size improves the accuracy of machine learning models. Accordingly, we find the feedback loop logic compelling. However, we show that the stock of available data produced by the feedback loop has a limited oracle size because of time dependency. Hence, despite the accelerated growth in the size of the data repository, we shouldn't expect a significant increase in created business value.  In other words, the feedback loop stalls in dynamically changing environments unless the firm offloads its less relevant data and focuses on the flow of data as the primary value driver. This finding supports the reported results in \cite{c13} and \cite{c5} since both search engine [\cite{c13}] and advertisement [\cite{c5}] businesses use time-sensitive data and hence, face significant time-dependency. 

Our work also adds to machine learning and economics literature in several ways. This paper adds to machine learning (particularly Natural Language Processing) literature by providing a different view of the domain/concept shift problem. Our method and analysis make machine learning researchers and practitioners better explain the tradeoffs and challenges that variation in training data has for their models. For example, our method allows them to realize how often they should retrain their models. As a result, they can formulate a better scaling/growth plan by adequately crafting their data management and resource prioritization strategy.  It is worth noting that in this paper, we only measure the predictive value of data. Hence, we don't talk about the value of data for inference. There is a subtle distinction between the two in the statistics literature.

We also contribute to the economics literature by providing a better understanding of data's time dependency and its implication on modeling data economy and the growth of digital firms. In the economics literature, the impact of data on economy and AI's widespread applications have been examined by several authors [e.g., \cite{c2},\cite{c3},\cite{c7},\cite{c10},\cite{c15},\cite{c16},\cite{c29},\cite{c30},\cite{c33},\cite{c38},\cite{c42},\cite{c44}]. In this literature, AI is a general-purpose technology that brings down the cost of prediction. Accuracy of the prediction increases with the size of training datasets, making a case for arguments on the importance of data in the growth of digital firms. Such arguments motivate research on how data influences firm dynamics [\cite{c20},\cite{c21}] and how it disproportionally benefits large firms [\cite{c8}], which then stimulates debates on the implication of AI and, more precisely, data on competition [\cite{c17},\cite{c18},\cite{c32},\cite{c34},\cite{c35},\cite{c39}]. Of course, the degree to which data impacts business value and hence the competition varies with the design parameters like the degree of personalization [\cite{c26},\cite{c40}] or the externalities between recommendation clusters [\cite{c6}]. Nevertheless, studying the effect of data's time dependency on competition remains crucial. Several lines of research [\cite{c11},\cite{c29},\cite{c32},\cite{c39},\cite{c43}] recruit the resource base view as a framework to explore how a firm can exploit data to create a sustainable competitive advantage. These researches mostly focus on the non-rivalry, exclusivity, or imitability of data. In our research, we argue that time-dependency plays a major role as well. We show that the business value doesn't solely scale with the size of available data, and even a dataset of infinite size may have a finite equivalent oracle size. As we discuss more in section 3, an immediate impact of this result is that we can't use regular discounting functions to model decay in the value of data. An adequate discounting model is a function of time and the size of the dataset. Our paper also distinguishes between the flow of data and stock of historical data in creating business value. Such distinction is also noted by \cite{c14} in their experiment in the context of online news. Our paper argues that small firms should focus on the stock of available data and gradually shift their focus to the flow of data as they grow over time.

In this paper, in the framework section, we explain our approach to the problem and clarify why we made particular choices. Then, in section 3, we introduce the effectiveness curve and explain value depreciation over time. We show the bounded size equivalence in this section. Section 4 investigates the effectiveness of datasets curated over time. These datasets are a combination of datasets samples from various times. We explain sequential offloading and suggest that old data may even put a firm in a disadvantaged position in businesses with high time dependency. Section 5 empirically measures the value depreciation for the next word prediction task. Finally, in the conclusion section, we wrap this paper with a discussion.

\section{Background and Framework}
In this section, we introduce our approach to the problem and explain the particular modeling choices we made. We start by describing the relevance loss of a dataset as a shift in the underlying data-generating distribution over time and dig into its possible cause. We argue that the relevance loss mostly stems from exceptional reasons that often cause a monotonic decrease in the value of data over time. We then provide a brief introduction to machine learning models and explain our focus on the data-generating probability distribution's maximum likelihood estimation. We introduce a decomposition of the MLE's objective function to lay the groundwork for the next section's propositions. 

\subsection{Change in Distribution}
Time-dependency of data occurs for many reasons. For example, it can happen if consumers taste or behavior changes over time. If we look at best seller music albums from the 80s and compare them with the best sellers in 2020, we can see the difference in taste. Another example is the continuous innovation in products and services space. Telegram is a perfect example of widely used innovation that is ancient nowadays due to the invention of other communication devices like hardline telephones. Soon, hardline telephones will be ancient too because of the introduction of voice over IP phones (VOIP) or cellphones. Given the rapid development of new technologies in this sector, if we tried to use consumer behavior data from the 1960s to predict how consumers will use the newest iPhone, such a task would be impossible and absurd. This fundamental shift in consumers behavior caused by innovation makes time-dependency a severe problem.

In the machine learning context, time-dependency is usually referred to as concept drift or non-stationarity. As an example, suppose we use letters written a hundred years ago to train a text auto-completion model for smartphones today. In that case, the users will be disappointed since the way we write and communicate has fundamentally changed over time which means that the model will suggest words or phrases that we no longer use. When this occurs, ML researchers try to use tools known as transfer learning to deal with this non-stationarity and combat time-dependency. These techniques usually describe the change in the data over time as a change in the data-generating probability distribution. Consequently, they either use the dataset's histogram to learn this change over time, or they assume a time-model for the change and accordingly adjust the ML models. In either case, dealing with the data generating distributions has its own problems. To name one, these solutions mostly approximate the change over time and hence, they still incur penalties for not being perfect. But most importantly, they assume that the set of elements $\Omega$ in the probability space $(\Omega,\sigma,P)$ is known in advance, which is a fairly big assumption and one that doesn't account for continuous innovation over time. Going back to our text auto-completion example, the data generating distribution changes over time in two ways. First, as time passes there is a lower probability for historical words and phrases to be used. Second, the environment (language in our example) allows for the birth of new elements (new phrases and words), which is equivalent to an increase in their frequency of use. An example of such words is \qq{covfefe} which President Trump used in a tweet and became viral. This word was never recorded in any dataset before the president coined it. This birth and death process of probability space elements, if not accounted for, is among the very reasons we see the depreciation in data value over time. Unfortunately, it is hard to predict and adjust for such changes in data-generating processes and despite the best effort of ML researchers, these transfer learning models are not perfect. In our paper, we assume that the prediction model is fixed, and it may account for transfer learning. Nevertheless, due to imperfection, achieved model accuracy changes depending on the time the training data was collected.

Similar to machine learning literature on transfer learning and non-stationary, we also observe time-dependency as an outcome of change in data-generating probability distributions. To compare distributions at different points in time, we create a universal set of elements. It is the union of all element sets across all time periods. For example, the word iPhone is created in the 2000s. In the language dataset from 1900, this element does not exist, and hence, it is not measurable. We create the universal element set by including this word. Then, for the dataset from the 1900, we should designate zero probability for its appearance. Still, instead of assigning this word a zero probability, we give it an infinitesimal value. This infinitesimal probability helps us use different functional forms like the log function without being worried about issues with functional domains.

Formalizing the assumptions we made so far, we assume that the prediction is for time 0 with a model trained on data from the past. The training data from the past is from $t$ periods prior to time 0 where $t \in \{0\} \bigcup \mathbb{R}^+$. For the sake of simplicity, we will say this historical data was sampled at time $t$. The element set at time $t$ is $\Omega_t$ for the probability space $(\Omega_t,\sigma_t, \widetilde{P}_t )$, where $\sigma_t$ is the sigma-field over $\Omega_t$ and $\widetilde{P}_t$ is the probability over the sigma-field. The universal probability space is then $(\Omega,\sigma,P_t)$, where $\Omega=\bigcup \Omega_t$ and $\delta=\sum_{\omega \in \Omega - \Omega_t} \delta_\omega$ and $\delta_\omega > 0$. As explained earlier, we prefer $\delta$ to be zero, but due to some regulatory conditions in the MLE's loss function, we assume $\delta$ is infinitesimal.
\[P_t (\omega)=\left\{\begin{matrix}
(1-\delta) \widetilde{P}_t  & \omega\in \Omega_t \\ \delta _\omega & \omega \in \Omega-\Omega_t
\end{matrix}\right. \]
With this change in measure, it is possible to compare datasets and define the shift in distribution. A change in distribution between the time $i$ and $j$ means $\exists \text{ } \omega \in \Omega \text{ s.t. } P_i(\omega) \neq P_j(\omega)$.

\subsection{Learning Data Distributions}
Machine learning aims to find the relation between inputs and outputs of an unknown system.  The unknown system is usually seen as a black box with little or no information about its function. The goal is then to observe examples of this unknown system's function and adjust a model's parameters accordingly so that the model can replicate the system's function as similarly as possible. These observed examples of systems' function are referred to as data. 

Putting this into mathematical semantics, given a dataset $D_{n,t}=\{(x_i,y_i )_t \}_{i=1}^n $, which is composed of $n$ input-output samples $d_i=(x_i,y_i )_t \in \Omega$ collected at time $t$, we want to find a model $m(d,\theta) \in M$ that describes the relationship between the input $x$ and the output $y$. Here, $\theta$ represents the model's parameters and $M$ is the set of all candidate models distinguished by the parameters $\theta$. Linear, logistic, and deep neural network compositional functions are examples of $m(d,\theta)$. Table 1 provides the functional forms for these three examples. In most machine learning cases, the goal is to make $m(d,\theta)$ as close as possible to the system's function by learning the parameters $\theta$. The notion of closeness depends on the problem formulation and the objective of the learning task.

\begin{table}[h!]
\caption{Examples of functional forms for famous ML models.}
\centering
\begin{tabular}{||c c||} 
 \hline
Case	& Functional form $m\left(d,\theta\right)$ where $d=(x,y)$ \\ [0.5ex] 
 \hline\hline
 Linear functional	& $y=\theta x$ \\
Logistic functional	 & $y=\frac{e^{\theta x}}{1+e^{\theta x}}$ \\
Simple Deep Learner with L layers and & \\non-linear functions $\gamma$ &	$y=\theta_L\gamma\left(\theta_{L-1}\gamma_{L-1}\left(\ldots\ \left({\ \theta}_2\gamma_2{(\theta}_1x\right)\ \right)\ldots\right))$ \\ [1ex] 
 \hline
\end{tabular}
\label{table:1}
\end{table}

In this paper, we restrict our theoretical analysis to the problem of learning the data-generating probability distribution. We do this because identifying this probability distribution is the fundamental problem in statistical learning theory. Once we know the probability distribution that characterizes the data-generation process, we are able to calculate any statistics of interest about the data, such as its expected value or variance. In general, all statistical models are a function of data distributions. Consequently, under specific regulatory conditions like continuity of models in the probability space, a sequence of distributions converging to the data-generating probability distribution also defines a converging sequence of the model to its converging value. This argument attests that learning the underlying distribution is the fundamental problem in machine learning. 

Finally, we choose the maximum likelihood estimator for learning the data-generating probability distribution since it is an unbiased and efficient estimator. Due to its efficiency, it is rational to prefer it over other unbiased estimators. Note that in this research, we are not concerned about time-complexity or other computational issues. Our goal is to get the most from a limited number of data points, and hence, we care about efficiency.

\subsection{Maximum Likelihood Estimation and Learning the Probability Distribution}
In the problem of learning a probability distribution, the unknown system is the distribution's functional form. The unknown distribution is defined over the set $\Omega$. The goal is to introduce an estimator $m(d,\theta)$ that converges to $P(\omega)$ for all $\omega \in \Omega$ as dataset size approaches infinity ($n\rightarrow \infty$).
The MLE's objective function for estimating the probability distribution, using the model $m(d,\theta)$ and the dataset $D_n=\{d_i \}_{i=1}^n$, has following form
\[
\theta_n= \argmax_{\theta} \sum_{i=1}^n \text{log} (m( d_i, \theta)) 
\]
By dividing the sum by the number of samples and multiplying it by -1, we reach the following equivalent minimization problem. The objective function denotes a loss function called empirical cross-entropy.
\[
\theta_n= \argmin_\theta -\frac{1}{n}\sum_{i=1}^n \log(m(d_i,\theta)) 
\]
Convergence to a local minimum happens as the size of the dataset, that is sampled independently and from an identical distribution, grows. For the sake of simplicity and for not dealing with issues of local optimums, we assume our optimization reaches the global optimum and $\lim_{n\rightarrow \infty} \theta_n=\theta^*$ where $m(d,\theta^* )=P(\omega)  $ $\forall \omega \in \Omega$. Of course, this is true with the assumption that $P \in M$ (The solution exists in the search domain). As explained earlier, in this paper we assume that algorithms are wisely chosen, and our goal is to see how they perform when the training and testing data are from different distributions. From the Central Limit Theorem, we can see the following approximation for the loss function's value.

\begin{theorem}
Assuming $E(\log m(x,\theta^*) )^2<\infty$, for a sufficiently large number of data points ($n>>1$), the loss function can be approximated with
\[\frac{-1}{n} \sum_{i=1}^n \log(m(x_i,\theta)) =H(P)+D( P||m(x,\theta)  )+ O(\frac{C_1}{\sqrt{n}})\mathcal{N}(0,1)\]
Where $C_1$ is a constant, and, is a function of $var(\log m(d,\theta^* ) )$. $H(P)$ is the Shannon entropy defined as $H(P)=-\sum_{\omega \in \Omega} P(\omega) \log (P(\omega))$ [\cite{c41}], and the summation is over the element set $\Omega$. $KL( P|| m(d,\theta_n ) )=\sum_{\omega \in \Omega} P(\omega)  \log( \frac{P(\omega)}{m(d,\theta_n) })$ is the Kullback-Leibler (KL) divergence [\cite{c31}] between the actual distribution $P(\omega)$ and the estimator/model $m(d,\theta)$. 

\end{theorem}

As the size of the dataset approaches infinity, the error term is getting smaller. Immediately from theorem 1, we can see that KL-divergence is the only component of the loss function that is a function of $\theta$ (Model). Hence, minimizing the loss function is equivalent to minimizing $KL( P||m(d,\theta))$. The property of KL-divergence is that it is always non-negative. Besides, it is equal to zero if and only if $P(d)=m(d,\theta)$ almost anywhere. With KL-divergence equal to zero, the only term remaining in the loss function is $H(P)$, which describes the system's entropy. The convergence speed of loss function to $H(P)$ as the function of dataset size is called the learning curve.

\subsection{Learning Curve}
Learning curves represent the expected value of the objective function (loss function in our case) versus the size of the dataset. The expected value is taken with respect to randomness in sampling or algorithm's initialization. In other words, fixing the size, if we sample the dataset and train the model infinite times and then take the average loss value, we reach the expected value of the loss function. Putting this in mathematical semantics, the learning curve is a function $G_t (n):\mathbb{R}^+\rightarrow \mathbb{R}$ that takes the size of a dataset as input and outputs the value we should expect for the loss function. For the problem of learning the data-generating probability distribution, this function shows how KL-divergence $KL( P||m(d,\theta))$ changes with the dataset's size. 

From theorem 1, with infinite sample size, we can see the loss function's convergences to the entropy of the data-generating distribution. Since the data-generating distribution changes over time, its entropy changes as well and hence, we added the subscript $t$ to $G_t (n)$ to capture this time-dependency. This function is monotonically decreasing and hence, invertible. Due to its asymptotic convergence to a bounded value $H(P_t)$, it has a convex form for large dataset sizes. We further assume that it is continuous and differentiable, meaning that ($\frac{\partial G_t(n)}{\partial n} <0$).

In practice, the learning curve is shown to be predictable for deep learning algorithms [\cite{c25}] and is composed of small data, power-law, and irreducible loss regions (Shown in Figure 1). In the small data region, the model is not scaling significantly with dataset size. The power-law region is where model performance scales with dataset size. In this region, for deep learners, the function $G_t (n)$ is believed [\cite{c25}] to have a power-law functional form. Lastly, in the irreducible loss region, the model's generalization loss value does not improve significantly. Between the regions, the power-law region is the one that we can see significant improvement in performance as we increase the dataset size. 

\begin{figure}[hbtp]
\centering
\includegraphics[height = 68 mm]{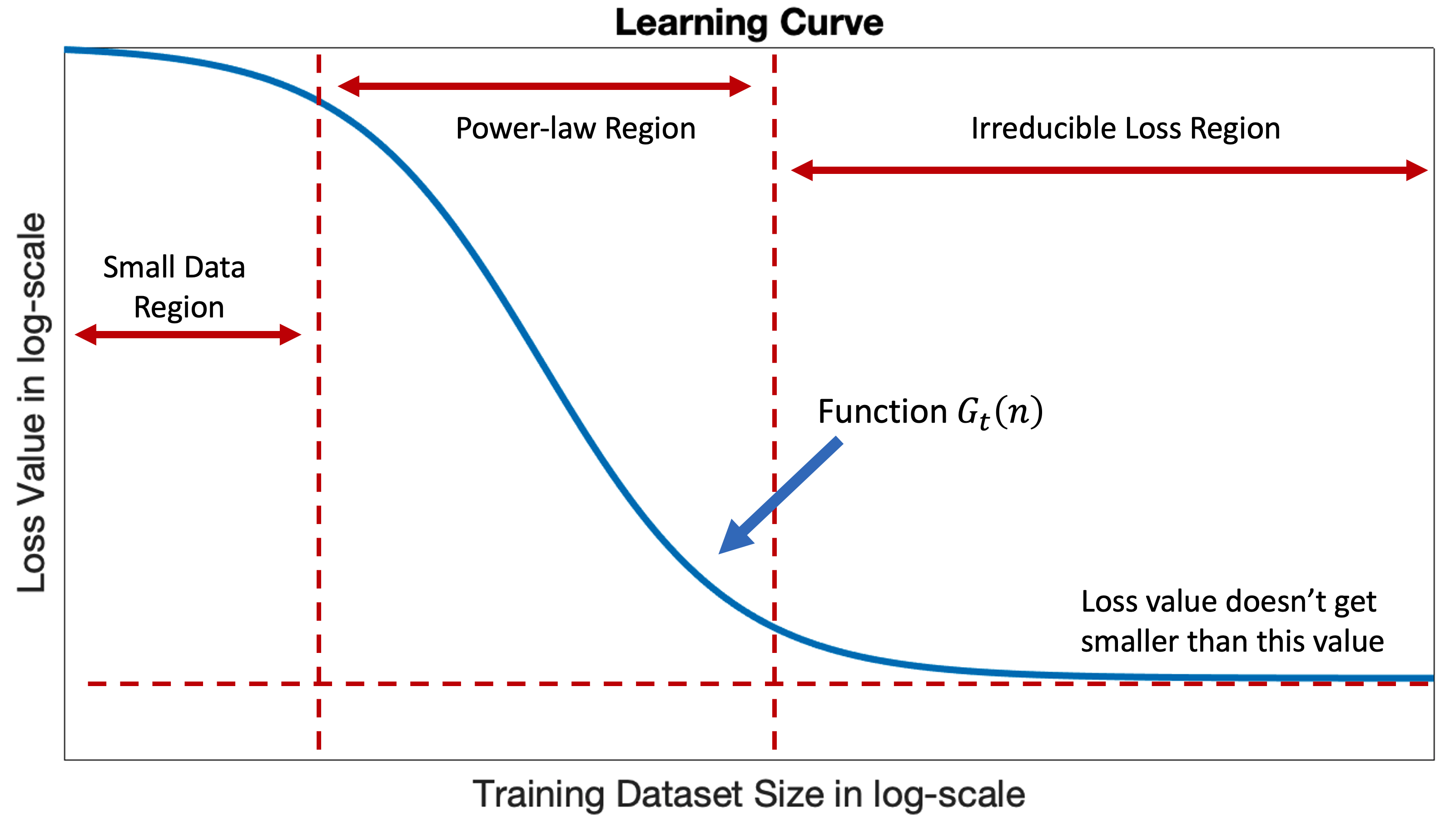}
\caption{Power-law learning curve.}
\end{figure}


\section{Effectiveness Curve and Value Depreciation }
To achieve our goal of measuring the depreciation of datasets' value over time, we need a mechanism to find the value of any given dataset at any given time. Unfortunately, value is subjective and dependent on the context, problem definition, and implementation. Therefore, we define a novel measure of datasets value that reports it as a function of the size. To explain this measure, we first define an oracle dataset that is sampled at time 0 from $P_0$. We use this oracle dataset as a base of comparison and say that the loss value that a model trained on this dataset achieves is our reference loss value and the best loss value we can achieve. We then compare the model's performance trained on the baseline dataset to the model's performance trained on oracle. We finally ask what amount of data from the oracle dataset allows our model to achieve the equivalent loss value that the model achieves after being trained on the baseline. The \qq{equivalent size} of the baseline dataset is the amount of data from the oracle dataset needed to train a model achieving the same loss values.  

Before formally defining the notion of equivalent size, a good starting point would be to compare a baseline dataset of infinite size and see how well-performing a model trained on this dataset is in predicting $P_0$ (the oracle). This is particularly important since we expect the infinite sized baseline dataset to be as valuable as possible in reaching the ultimate algorithm's performance. In a way, this gives us clues on whether the infinite baseline dataset can scale created value. Proposition 1 investigates this ultimate scaling behavior.

\begin{proposition}
Assuming $P_t\left(\omega\right)\neq P_0\left(\omega\right)$, a model trained on a dataset of infinite size from the wrong distribution $P_t\left(\omega\right)$ has limited predicting power at time 0, and in probability, a model trained on a dataset of bounded size from the right distribution $P_0\left(\omega\right)$ can reach the same loss value.
\end{proposition}

The argument in proposition 1 is that in the training phase, due to the change in the probability distribution, $m(d,\theta)$ convergences to the wrong distribution $P_t\left(\omega\right)$. Therefore, MLE's loss value has an additional term $KL\left(P_0||P_t\right)$ besides the Shannon entropy $H(P_0)$. It is as if we used a dataset of bounded size from $P_0(\omega)$, and due to its limited size, we did not reach the minimum loss value possible. The minimum loss value possible is reached when MLE's loss function is equal to $H(P_0)$. 

Intuitively, this proposition tells us that created value in a machine learning-based product or service scales differently with the size of datasets compared with the way tangible assets scale the value. For example, suppose we have a company that produces apple juice boxes. Apples that are stored for a little while longer in inventory will dry out and produce less juice compared to fresh apples. Nevertheless, if we scale the number of older apples in the juice production business, the number of potential juice boxes scales with it. In contrast, when it comes to time-dependent data, increasing the number of older data points doesn't necessarily drive an increase in the model's accuracy and, thereby, the created business value. This is because an oracle dataset of bounded size can produce a loss value equal to that of an infinite amount of older (less relevant) baseline data.

This proposition has implications in economic modeling, academic antitrust debates, and data management strategy for practitioners. Proposition 1 states that curating massive datasets over time does not create a significant barrier to entry of a competitor if the data-generating distribution changes. In our interviews with practitioners, we always found them hopeful that increasing dataset size can compensate for the shortcomings in scaling. On the regulatory side, also there are debates on the role of super large datasets and if they create a barrier to entry advantage for big tech companies. This proposition states that scaling of value is different from what we previously knew and hence, the role of time-dependency should be accounted for. This proposition has also implications for economic modeling. The immediate implication is the way we should model the decline in the value of data. Specially, when modelers want to treat data as an asset, they should be aware of the way they account for the value decline. The usual approach to account for the value decline is to use a time-dependent discounting function like an exponential decay $e^{-ct}$ to be multiplied by the accumulated capital at time $t$. This proposition states that such discounting functions should be a function of accumulated capital as well as the sampling time since there is not a multiplicative discounting function that is multiplied in infinity and results in a finite value. Hence, the discounting function should be a function of both size and the time.

As explained earlier, something lacking from proposition 1 is that it talks about loss value, which is not very informative in making comparisons. It is not informative because we do not know how to interpret the excess loss value term $KL\left(P_0||P_t\right)$. We just know that it is positive, and therefore, the loss value should be bigger than the one for the oracle dataset. To solve this issue, we use the learning curve inverse function to translate the loss function back into an \qq{equivalent} dataset size. Dataset sizes are easy to understand and compare. 

Recall that learning curve at time zero $G_0(n)$ is a monotone function and therefore has an inverse. Using the inverse of the learning curve $G_0^{-1}(.)$, we can find the expected size of a dataset from time zero (the oracle dataset) with an equivalent MLE loss value. Briefly, what we do to form the equivalent size is to first train a model on data sampled from $P_t(\omega)$. Then, we use the trained model to find the loss value on the data that has been sampled from $P_0(\omega)$. Finally, we use the function $G_0^{-1}(.)$ to see what size of the data from $P_0(\omega)$ can generate a similar loss values. This is the basis for our definition of equivalent size.

\begin{definition}
Dataset $D_{n,t}$ has the equivalent size ${\bar{n}}_{D_{n,t}}$ at time 0:
\[{\bar{n}}_{D_{n,t}}=E_{\theta_{n,t}}\left(G_0^{-1}\left({-E}_{P_0}\left(\log{\ m\left(d,\theta_{n,t}\right)}\right)\ \right)\right)\]
Where $\theta_{n,t}$ is the solution to: 
\[\theta_{n,t}={\rm argmin}_\theta{-\frac{1}{|D_{n,t}|}\sum_{d\in D_{n,t}}\log{\left(m\left(d,\theta\right)\right)}}\]
\end{definition}
In this definition, there exist two expectations. The expectation inside of $G_0^{-1}(.)$ measures the expected model's loss over the test set. The second one is the outer expectation and calculates the expectation with respect to randomness in the algorithm's initialization and steps. In practice, we can approximate the outer expectation by solving for $\theta_{n,t}$ multiple times. Using averaging limit, we can calculate the equivalence empirically in the following way
\[\lim_{k\rightarrow\infty}{\frac{1}{k}\sum_{j=1}^{k}\left(G_0^{-1}\left(\lim_{l\rightarrow\infty\ }{-\frac{1}{l}\sum_{i=1}^{l}\log{\left(m\left(d_i,\theta_{n,t}^{\left(j\right)}\right)\right)}}\right)\right)}\]
Where $d_i\sim P_0(\omega)$ and the outer sum is over multiple runs of the algorithm. For a fairly large number of testing data points, the inner expectation converges. Using theorem 1 to simplify the definition further, we have
\[{\bar{n}}_{D_{n,t}}=E\left(G_0^{-1}\left(H\left(P_0\right)+KL\left(P_0||m\left(d,\theta_{n,t}\right)\right)\ \right)\right) \]
Letting $n\rightarrow\infty$ eliminates algorithms' initialization issues as well as other types of randomness and hence, $m\left(d,\theta_{n,t}\right)\rightarrow P_t\left(\omega\right)$.  Therefore, in the limit
\[{\bar{n}}_{D_{\infty,t}}=G_0^{-1}\left(H\left(P_0\right)+KL\left(P_0||P_t\right)\ \right)\]
It is in agreement with proposition 1 where it argues that ${\bar{n}}_{D_{\infty,t}}<\infty\text{ if } P_0\left(\omega\right)\neq P_t(\omega)$. Because $G_0^{-1}(H\left(P_0\right)+KL\left(P_0\middle|\left|P_t\right)\right)<G_0^{-1}\left(H\left(P_0\right)\right)=\infty$.

 Notice that the equivalent size is a function of the algorithm as well as the dataset itself. Dependence on the algorithm is recognized through the inverse function $G_0^{-1}(.)$. It means that the algorithm's power in scaling with dataset size shapes the effectiveness of a dataset. The following example makes it clear. Suppose we have a very large dataset, but we do not use it to train a model. In that case, the sampling time is not essential and, regardless of time, the dataset is as effective as not having it in the first place ($n=0$). On the other hand, if the algorithm scales in a faster pace (in the number of data points), a small dataset from $P_0\left(\omega\right)$ can reach $H\left(P_0\right)+KL\left(P_0||P_t\right)$ with a smaller number of  data points which means ${\bar{n}}_{D_{\infty,t}}$ is indeed small.

\begin{definition} Effectiveness of dataset $D_{n,t}$ is defined as $E_{D_{n,t}}=\ \frac{{\bar{n}}_{D_{n,t}}}{n}$. 
\end{definition}

Intuitively the effectiveness should be always between zero and one, i.e., $E_{D_{n,t}}\in[0,1]$. $E_{D_{n,t}}=1$ means that the given dataset's value is equal to the value of the oracle dataset. Note that $E_{D_{n,t}}$, by definition, can't be more than 1. $E_{D_{n,t}}=0$ means that data is worthless compared to the oracle dataset and the prediction power of a model trained on this dataset is equivalent to a uniformly random guessing of output values. The value of $E_{D_{n,t}}\in[0,1)$ signals that the equivalent size is less that the actual size of a dataset. It is as if dataset perishes over time. The more perishable the data, the less its effectiveness over time. For example, if the effectiveness is equal to 0.8, we say that the dataset lost 20\% of its effective size. 

Proposition 1 argues that effectiveness $E_{D_{\infty,t}}=0$ if $P_0\left(\omega\right)\neq P_t(\omega)$. It is because ${\bar{n}}_{D_{n,t}}$ remains bounded and therefore, $\lim_{n\rightarrow\infty}{\frac{{\bar{n}}_{D_{n,t}}}{n}}=0$.

\begin{definition}
Substitution curve is a function $f_n\left(t_1,t_2\right):\mathbb{R}^2\rightarrow\mathbb{R}$ and is defined as 
\[f_n\left(t_1,t_2\right)=\frac{{\bar{n}}_{{D_{n,t}}_1}}{{\bar{n}}_{D_{n,t_2}}}\]
\end{definition}
It shows how well we will be off in terms of effectiveness if we substitute a dataset of size $n$ from time $t_2$ with a dataset of the same size that has been sampled at time $t_1$. Note that choosing $t_2=0$ brings us back to the definition of effectiveness. Using theorem 1, the substitution curve has following formulation
\[
f_n\left(t_1,t_2\right)=\frac{{\bar{n}}_{{D_{n,t}}_1}}{{\bar{n}}_{D_{n,t_2}}}=\frac{E\left(G_0^{-1}\left(H\left(P_0\right)+KL\left(P_0||m\left(d,\theta_{n,t_1}\right)\right)\ \right)\right)}{E\left(G_0^{-1}\left(H\left(P_0\right)+KL\left(P_0||m\left(d,\theta_{n,t_2}\right)\right)\ \right)\right)}\]

\begin{theorem}
Substitution curve has the following properties. 
\begin{itemize}
\item[a)]	 It is non-negative and bounded. 
\item[b)] It is a monotonic function of n.
\item[c)] It is converging to a substitution frontier
\end{itemize}
\[\lim_{n\rightarrow\infty}{f_n\left(t_1,t_2\right)}=\frac{{\bar{n}}_{{D_{\infty,t}}_1}}{{\bar{n}}_{D_{\infty,t_2}}}=\frac{G_0^{-1}\left(H\left(P_0\right)+KL\left(P_0||P_{t_1}\right)\ \right)}{G_0^{-1}\left(H\left(P_0\right)+KL\left(P_0||P_{t_2}\right)\ \right)}\]
\end{theorem}

Nonnegativity and boundedness are immediate. It is nonnegative because function $G_0^{-1}$ is non-negative by definition. Boundedness is also immediate from proposition 1, because for $i\in\left\{1,2\right\}$ and $t_i\neq0$, $0<{\bar{n}}_{D_{n,t_i}}<\infty$ for all $n$.

The substation curve is an important definition in this paper. It is a building block for the argument we make in the next section on the effectiveness of datasets gathered over a long period of time. The concept will be used in the sequential offloading algorithm defined in the next session.  

Figure 2 depicts examples of substitution curves $f_n\left(t,1\right)$ assuming a monotonic decline in the value of data over time. Each curve represents the substitution gain over time t when the substitution time is fixed at $\left(t_1,t_2\right)=(t,1)$. Blue curve is the frontier trajectory $f_\infty(t,1)$ described in Theorem 2c. This figure pictorially shows the substitution function's monotonicity on n and its convergence to the frontier. As apparent in this Figure, we do not gain much in substituting data from different times for very small dataset sizes. It is because small datasets do not provide significant scaling in performance, and hence, it does not matter when they were sampled. This behavior is mostly seen in the small data region of the learning curve. For medium dataset sizes, when we are in the learning curve's power-law region, we gradually see significant gains in substituting datasets from different times. Increasing dataset size in the power-law region brings us to the medium-high dataset size regime. This region will be used in our experiments (In later sections) to measure perishability. Finally, the infinite dataset size speaks of the irreducible loss region and the highest sensitivity to substitution.

In Appendix B, building further on our observation, we empirically measure the substitution curve for our experiment in this paper and show that gain/loss increases in $n$ for $t_1>t_2$ and decreases for $t_1<t_2$.  In other words, the substitution curves become sharper and the gains/loss in substituting data from various time increase. Proposition 2 formalizes this idea. To prove this proposition, we have two additional assumption. First, we assume that the monotonicity result proved in theorem 2 (b) is valid for all dataset sizes meaning that $f_n(t_1,t_2)$ is monotonic for all $n$. Second, we assume that $f_1\left(t_1,t_2\right)=1$ for all $t_1,t_2\in\mathbb{R}^+\cup{0}$. It is intuitive since, in our model, all elements have non-zero probability and hence, one data point carries in expectation same amount of information regardless of when it was sampled.
\begin{proposition}
The substitution gain function becomes sharper with the increase in the size of the baseline dataset ($n$), meaning that the gain/loss increases as the baseline dataset size increases. Mathematically, for all time $t_1,t_2\in\mathbb{R}^+\cup\left\{0\right\}$ and sizes $n_1,n_2\in\mathbb{N}$ with  $n_1<n_2$:
\[\left\{\begin{matrix}
f_{n_1}(t_1,t_2) \leq  f_{n_2}(t_1,t_2) & \text{   when  } &  f_{n_1}(t_1,t_2) \geq 1\\
f_{n_1}(t_1,t_2)> f_{n_2}(t_1,t_2) & \text{   when  } &  f_{n_1}(t_1,t_2) < 1
\end{matrix}\right. \]
\end{proposition}

\begin{figure}[hbtp]
\centering
\includegraphics[height = 75mm]{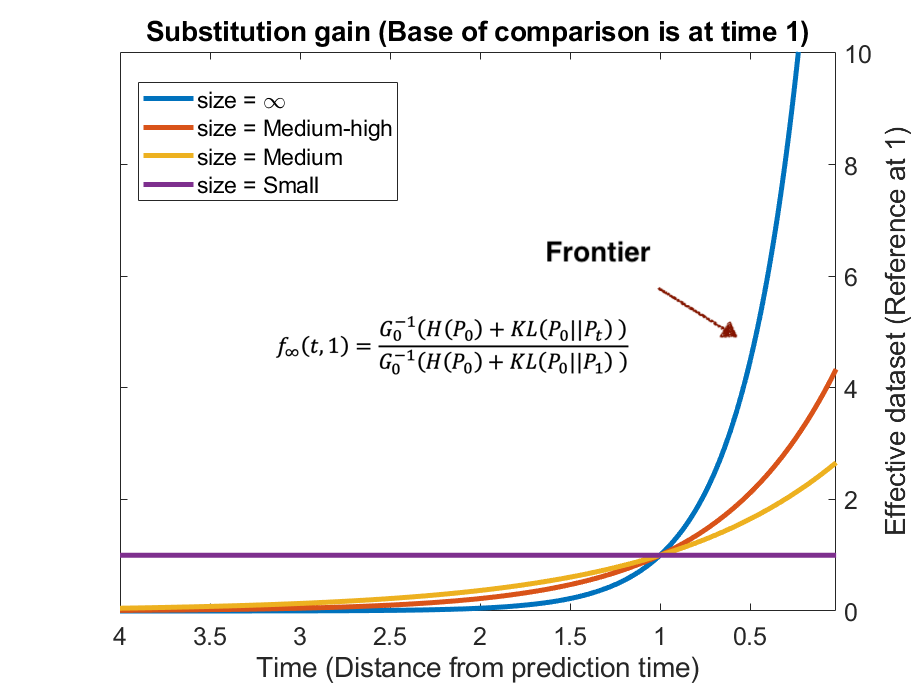}
\caption{Substitution curves for different sizes of datasets. The frontier is marked as blue. It shows the maximum depreciation in substituting datasets of time 1 with a dataset of any other time.}
\end{figure}

This proposition is particularly crucial for our discussion in section 4. We use it to prove that flow of data becomes the main value driver when a firm grows, for example, in the user base. Hence, firms should retrain their models more frequently when the amount of data they collect in a given time ($n$ in this case) grows. 

\section{Datasets Collected Over Time}
So far, we have studied the effectiveness of datasets that have been sampled at a given time $t$. Nevertheless, most datasets are collected over time, and there is a need to study their effectiveness and compare their values. This is particularly important since we want to study the historical value of data. In that pursuit, we compare two datasets that have been sampled over time; A baseline dataset that spans over a longer period and a subset of that dataset which has a smaller size and contains only recent data (most relevant data in case of semi-monotonic decline in value of data over time). When the value of smaller dataset is more than the value of larger dataset, we conclude that using historical data in the training set increases the loss value and may put a company in a disadvantageous position. In section 4.1, we use this idea and introduce the sequential offloading algorithm. This algorithm exploits the tradeoff between the size and the relevance of datasets gathered over time. It removes less relevant data from the training dataset, meaning we lose size, and hopes that increasing the relevance counterbalances the loss in size.

We use the machinery we developed so far, with slight modification, to compare the values of datasets sampled over time. In doing so, we first show that the accuracy score (loss value) of models trained on such datasets are equivalent to the accuracy score (loss value) of models trained on a dataset of similar size that has been sampled at a time $t^\ast$, i.e., there exist a time $t^\ast$ such that a model trained on a dataset from this time has an equivalent loss value to the same model that is trained on the dataset gathered over time. $t^\ast$ is called the equivalent time. Subsequently, we calculate the equivalent time for these datasets (that have been sampled over time) and use the substitution curve to compare their values.

Notation wise, we show datasets of size $n$ that are collected over a period $\left[t_1,t_2\right]$ with $D_{n,\left[t_1,t_2\right],\lambda_t}$. $\lambda_t$ is called the sampling density function and shows the proportion of samples that have been collected at time t. For the sake of simplicity, we assume $t_1 =0$ and focus on datasets of the form $D_{n,\left[0,t\right],\lambda_t}$. Mainly because it is easier to turn a bigger period of time into smaller periods with the sampling function $\lambda_t$.  For example, the dataset $D_{n,\left[t_1,t_2\right],{\hat{\lambda}}_t}$ is equivalent to dataset $D_{n,\left[0,t_2\right],\lambda_t}$ with $\lambda_t$ equal to
\[
\lambda_t=\left\{\begin{matrix}
 0  & t<t_1 \\
 \hat{\lambda}_t & \text{    } t_1\leq t \leq t_2
\end{matrix}\right.
\]
Although the dataset $D_{n,\left[t_1,t_2\right],\lambda_t}$ is sampled from various time with different generating probability distribution functions, it still has a \qq{Net Distribution} that can be used to measure its effectiveness. Lemma 1 presents the net distribution for $D_{n,\left[0,t\right],\lambda_t}$. It states that the net distribution is a convex combination of all distribution from time 0 to t with weights $\lambda_t\in\left[0,1\right]\ \&\ \int_{0}^{t}{\lambda_sds=1}$ .

\begin{lemma}
Net distribution of dataset $D_{n,\left[0,t\right],\lambda_t}$ is equal to 
\[P_{\left[0,t\right],\lambda_t}(\omega)=\int_{0}^{t}{P_s\left(\omega\right)\lambda_sds}\]
\end{lemma}

As this lemma states, the net distribution is not necessarily equal to $P_0$. Therefore, using proposition 1, we can still argue that datasets curated over a period of time have limited relevance. 
\begin{corollary}
 Datasets that are collected over time from dynamically changing environments have finite effectiveness and value. Hence, the growth between the accuracy of machine learning models and the stock of available data (Known as the data network effect) stalls.
\end{corollary}

Data network effect (The growth cycle between the stock of data and the accuracy of models) stalls when we have time-dependency. As stated in Corollary 1, the accuracy of machine learning models and the value they create doesn't simply scale with the stock of available data. As a result, a firm should either incorporate the time dimension into their models or alternate datasets to improve its relevance. Incorporating time in the models is not easy and often impossible, mainly when dealing with innovation over time. For example, we can't tell what kind of medical innovations we should expect in the next couple of years or what news we should read in the papers next week. Besides, we already assumed that firms already choose best models, and hence, we expected them to incorporate time if possible. Therefore, as the next step in this paper, we focus on methods to alternate the dataset's composition and improve their relevance. Alternating datasets should be in the direction of improving their relevance, and hence we first need a method to compare the value of dataset pre and post alteration.  Proposition 3, using the net distribution of datasets, maps the accuracy score of the dataset gathered over time to the accuracy score of a dataset that has been sampled in a given time. Hence, we can compare two datasets using the substitution curve method we developed in section 3.

\begin{proposition}
There exists an equivalent time $t^\ast\in[0,t]$ such that the dataset $D_{n,\left[0,t\right],\lambda_t}$ provides an equivalent loss value to the dataset $D_{n,t^\ast}$ i.e. $\overline{n}_{D_{n,[0,t]}}=\overline{n}_{D_{n,t^\ast}}$. The solution is unique when decline in value of data is monotonic.
\end{proposition}

Proposition 3 is the key to understanding the next subsection on sequential offloading. As much as it is important to understand what it says, it is also important to realize what it does not. It does not say that the \qq{Net distribution} is equal to $P_{t^\ast}$. Net distribution is a combination of many distributions, including $P_{t^\ast}$, and therefore, it is not necessarily equal to $P_{t^\ast}$. Instead, Proposition 3 suggests that $P_{\left[0,t\right],\lambda_t}$ and $P_{t^\ast}$ are in a way that they make equal KL divergences with $P_0$, i.e. $KL\left(P_0||P_{\left[0,t\right],\lambda_t}\right)=KL\left(P_0||P_{t^\ast}\right)$ . Consequently, they produce equivalent MLE loss value, which means ${\bar{n}}_{D_{n,\left[0,t\right],\lambda_t}}={\bar{n}}_{D_{n,t^\ast}}$. 

Note that having $t^\ast$ between zero and $t$ is important in this proposition. The emphasis is on the fact that the period $[0,t]$ starts from time 0. Even if the dataset has been sampled from $\left[t_1,t\right]$, still, $t^\ast\in[0,t]$. It is because, for the dataset ($D_{n,\left[t_1,t\right],\lambda_t}$) where $0<t_1<t$, there might exist a sampling density $\lambda_t$ such that it makes the Net distribution $P_{\left[0,t\right],\lambda_t}\left(\omega\right)=P_0(\omega)$ for all $\omega\in\Omega$, i.e. $t^\ast=0\notin[t_1,t]$.

The most exciting thing about this theorem is that  $t^\ast<t$. If we deliberately delete the portion $[t_1,t]$ from the dataset where $t_1<t^\ast$, despite losing size, the remaining dataset ($D_{n_1,\left[0,t_1\right],\lambda_t}$) will have a new equivalent time $t^{\ast\ast}$ which is $t^{\ast\ast}<t^\ast$. In other words, datasets gain relevance. Altering a dataset composition by deleting the portion $[t_1,t]$ is what we investigate next as sequential offloading algorithm.

\subsection{Sequential Offloading}
The idea of sequential offloading is founded in increasing the value of a dataset by reducing its size. It looks to be counter-intuitive, but in a time-dependent context, data perish quickly, and it may be beneficial to discard useless information. Clearly, deleting old data means loss of dataset size. Nevertheless, gaining relevance may offset the loss of dataset size, and deletion likely improves the overall effectiveness.

The idea is centered around Proposition 3 and theorem 2. Proposition 3 states that for a dataset $D_{n,\left[0,t\right],\lambda_t}$ there exist a time $t^\ast\in[0,t]$ such that ${\bar{n}}_{D_{n,\left[0,t\right],\lambda_t}}={\bar{n}}_{D_{n,t^\ast}}$. By deleting data $[t^\ast,t]$ from the dataset, we end up with a smaller size $n_0$, but the equivalent time shifts from $t^\ast$ to $t^{\ast\ast}\in[0,t^\ast]$, which is more relevant. If the substitution gain is higher than the lost size due to deletion, it means we gained from deletion i.e.
\[f_{n-n_0}\left(t^{\ast\ast},t^\ast\right)>\frac{n}{n-n_0}\Rightarrow\ {\bar{n}}_{D_{n,\left[0,t\right],\lambda_t}}<{\bar{n}}_{D_{n-n_0,\left[0,t^\ast\right],\lambda_t}}\]
Where $n_0$ is the size that has been deleted from the dataset. 
Algorithm 1 Formalizes the sequential offloading. This algorithm stops when there is no gain in deleting old data. It also opens a philosophical question on what a successful iteration means for the data. A successful iteration means $\overline{n}_{D_{n,[0,t]}}<\overline{n}_{D_{n-n_0,[0,t^\ast]}}$ and hence, there is positive improvement upon losing a portion of data. Therefore, as the following corollary states, including less relevant data (older data in our case) actually did put us in a disadvantageous position.

\begin{corollary} Including older (less relevant) data in the training set may put a firm in a disadvantageous competitive position. \end{corollary}

\begin{algorithm}
\caption{Sequential offloading algorithm}\label{alg:cap}
\begin{algorithmic}
\Require Dataset $D_{n,\left[0,t\right],\lambda_t}$, substitution gain function $f_n\left(t_1,t_2\right)$
\State $i \gets 1$
\State $t^{(0)}\gets t$
\State $n^{(0)}\gets n$ 
\While{(Gain is possible)}
\State Find $t^\ast$ as explained in theorem 2 and call $t^{(i)}$
\State $n^{(i)} \gets n^{(i-1)}\times \int_{t^{(i)}}^{t^{(i-1)}}\lambda_tdt$
\State Delete sampled data $[t^{(i)},t^{(i-1)}]$ from $D_{n^{(i-1)},\left[0,t^{(i-1)}\right],\lambda_t}$ and call it $D_{n^{(i)},\left[0,t^{(i)}\right],\lambda_t}$
\If{ $\left(f_{n^{(i)}-n^{(i-1)}}\left(t^{(i)},t^{(i-1)}\right)>\frac{n^{(i-1)}}{n^{(i)}-n^{(i-1)}}\right)$}
 \State		Gain \underline{is} possible
              \State $i \gets i+1$
               \Else
               \State Gain \underline{is not} possible
\EndIf

\EndWhile
\end{algorithmic}
\end{algorithm}

\subsection{Data as a Driver of Growth}
A successful offloading iteration means that older (less relevant) data is lowering the modeling accuracy score. Hence, it is optimal for a firm to retrain its models more frequently with relevant datasets. The retraining frequency is determined by the sharpness of the substitution function. From Proposition 2, we know that increasing the flow of data which is the rate of accumulating new information increases the gain/loss value of the substitution function and hence, changes the retraining frequency. Since the increase in flow is inevitable due to growth in the user-base or engagement, it is of paramount importance to understand the optimal growth strategy for a firm that derives value from data. In Proposition 4, we show that a sharper substitution gain function (which is the result of increase in flow) brings the equivalent time closer to the prediction time.

\begin{proposition}
Increases in the gain/loss value of the substitution function, as a result of multiplying the data flow rate, brings the equivalent time closer to the prediction time. 
\end{proposition}

We now cite Proposition 2 and argue that an multiplying the flow of data, with a constant $\alpha>1$, makes the offloading more likely, and hence, as stated in Proposition 4, it brings the equivalent time closer to the prediction time. Bringing the equivalent time closer to prediction time means that we rely on the flow of data to create value. Consequently, we conclude that when a firm grows, which leads to inevitable growth in the flow of data, it should shift its focus from the stock of available data to the flow of data as the primary value driver.

\subsection{Weight-Adjusted Datasets}
In previous subsections, we compared the values of two datasets, a baseline dataset that is collected over a period of time and a subset of this dataset that only contains recent data. In this comparison, we argued that there are cases where the subset of the baseline dataset has a higher value for a business application since the model trained on this subset has a higher accuracy score. Consequently, we argued that including older (less relevant) data in the training set might put a firm in a disadvantageous competitive position. We made this comparison to contest a widely known idea that having more data is always better, and as a result, companies who were collecting data earlier than others are in a better competitive position solely because of the data they own. 

Still, despite feasibility and ease of implementation, deleting older data from the dataset is a sub-optimal action from the implementation perspective. A wiser choice is to use a weight-adjusted version of the baseline dataset. The sequential offloading algorithm is also weight-adjusted since it puts zero weight on older data and full weight on recently collected data. However, this zero-one weighting is not necessarily optimal despite being more advantageous than the baseline dataset.

The main challenge is to find the optimal weights. It is challenging because of two reasons. First, as proposition 4 suggests, the value of data sampled from various times changes with the size of the dataset or the flow of data. Therefore, companies should continuously reevaluate and adjust weights over time since their dataset size or user-base changes. The second challenge is that companies should use the whole dataset to evaluate the weights, which is time-consuming and expensive to implement. Because, as proposition 2 suggests, the importance or value of recent data increases with the size of the subset, meaning that for larger subsets, the optimal weights of recent data will be more significant. Therefore, using a subset of the baseline dataset to evaluate the weights (Which is a standard solution to calibrate parameters in practice) miscalculates the weights and undervalues the importance of recent data. Because of these two points, i.e., continuously evaluating the optimal weights using the entire dataset a firm owns, we argue that it might be tractable to discard older data than repeating complicated repetition of the learning process. 

There is a question on the transferability of wights between firms. In other words, since the cause of perishability mentioned in this paper are at the market level, it makes sense that one firm calculates and sells the optimal weights to other firms. In that case, it is essential to know that weights depend on the company size, and the optimal weight that is useful for a big firm is not optimal for a smaller firm. 
\section{Experimental Design}
Our goal in this section is to measure effectiveness and thereby perishability of datasets empirically. In other words, after training an algorithm with data that has been sampled on one stationary period, we measure its equivalent oracle size at other periods. As it is shown later in this section, we observe a semi-monotonic decline in the value of a dataset. We expected the decline to be monotonic since datasets from the past lose relevance monotonically due to continuous innovation over time. However, the equivalent size has slight periodicity in the measurements due to seasonality in certain topics like fashion or other periodic data generating sources. Note that the overall decline still looks monotonic.

We make the measurements in the language modeling contexts.  Mainly because language modeling datasets tend to be the largest and most easily collected. Datasets are easily collected since language modeling tasks are most often unsupervised; For example, in the next word prediction task, the model predicts the next word in a given sentence, and hence, any book, magazine or text source could be a potential data source. Furthermore, the language modeling is currently used as a common pre-training objective for many other language tasks [\cite{c45}] making our measurements relevant to a wide variety of applications. In this section, we first explain data and how we process it for the task. Then, we explain the algorithm and model architecture, and lastly, we present the measurements.

\subsection{Data Collection and Processing}
Our challenge is to find a large enough dataset that has been collected over a long period of time. It is because text processing algorithms require large training set sizes to have significant improvement in quality. In addition, we need this dataset to be sampled over a long period of time to let us make an observable perishability measurement. From a technical standpoint, the dataset must be large enough to reliably measure the power-law portion of the learning curves associated with each time period. Thus, the dataset must span roughly two orders of magnitude in size larger than the smallest dataset in the power-law region. Prior results [\cite{c25}] show that, for language modeling, the smallest such dataset is at most 1 million words. Consequently, the dataset should contain roughly 10-100 million words per time period. 

We choose the Reddit post dataset as it fits to our needs. This data was collected and used in [\cite{c19}]. It is a collection of posts and comments from the years 2006 to 2018 and was scraped from Reddit between September 2016 and July 2018. We preprocessed the dataset to create flat text files with the following format:
{\begin{addmargin}[4em]{1em}
{\color{SeaGreen} Title (}6{\color{SeaGreen}):} What was the biggest scandal in your school?\\
{\color{SeaGreen}Text:}\\
{\color{SeaGreen}Comment (}4{\color{SeaGreen}):} Vampires. This was almost 6 years ago now at my high school, but vampires. Do a quick...\\
{\color{SeaGreen}Comment (}3{\color{SeaGreen}):} Not sure if I'd call it a \qq{scandal}, but when I was in college...\\
{\color{SeaGreen}Comment (}2{\color{SeaGreen}):} Freshman year a friend of mine found a paper bag at the bus stop full of money - and it...
\end{addmargin}}

'Title' is the title that the author specified when posting the submission, and 'Text' is an optional field of body text associated with the post. After the post, each line is a comment from other users designated by 'Comment'. Comments only contain text. The values in parenthesis are submission or comment scores based on upvotes or downvotes given to each by users. We filtered out comments and posts with scores less than 2. 

\begin{figure}[hbtp]
\centering
\includegraphics[scale=1]{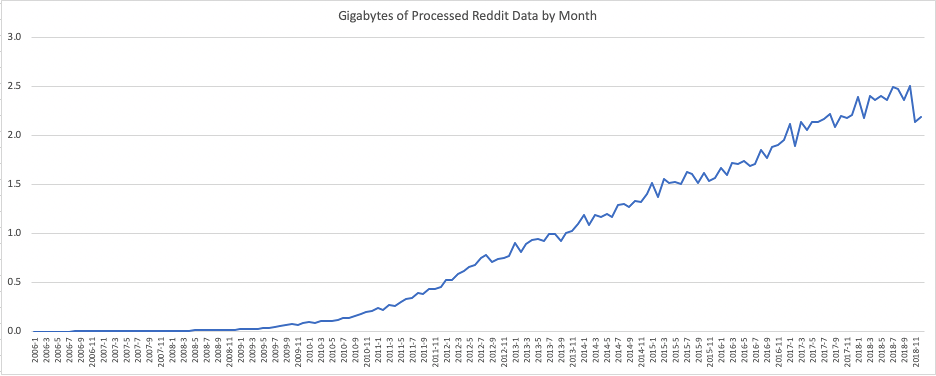}
\caption{Size of datasets processed for each month. For example, for July 2013, 1 Gigabyte of text data is processed. This is not a cumulative dataset size. The growth in the size shows the growth in the number of topics discussed, the number of users as well as their engagement. }
\end{figure}

To evaluate how data distributions and value shifts over time, we split the dataset into chunks based on the time stamp of the submissions and comments. We aim for 100 million words per time period, so we group data until each split is at least that large. Specially, we group posts and comments into the following periods: the years 2006-2009, January-June 2010, July-December 2010, January-March 2011, April-June 2011, July-September 2011, October-December 2011, and then monthly for the years 2012-2018. Earlier years of Reddit dataset have less data because the platform was becoming established and growing, so we had to group more extended periods together. Figure 3 shows the amount of data we processed each month.

Finally, we subdivide the data from each time period to form a standard machine learning training and testing setup for collecting learning curves. First, we randomly sample and split the posts (and their comments) into training, development/validation, and test/evaluation subsets. The development and test sets are at least 2 million words each. The development set is used to validate that the model is learning to generalize during training and to early-stop training when the model performs the best on the development set. The test set is used after training to evaluate how well the training is done. We use these test sets to cross-evaluate models trained on data from other periods. The model never trains on these subsets.

After splitting out the development and test sets, we randomly shuffled the remaining data as the full training set for the time period. We subdivide this training set into chunks of exponentially increasing size by factors of 2. Empirically, we find that datasets of size 1.25 million words are large enough to be in the power-law portion of learning curves, so we break the training set into successively overlapping subsets of size 40 million, 20 million, 10 million, 5 million, 2.5 million, 1.25 million words by taking the first half of the prior subset. We train separate models on each training subset to collect how models generalize as they are allowed to train with increasing dataset size. The resulting data size-generalizability curves are learning curves for the time period. 

\subsection{Model Architecture and Training Process}
We chose to train current state-of-the-art language models on the data to collect their generalization error and learning curves. Specifically, we train GPT-2, the Generative Pre-Training transformer-based model from OpenAI [\cite{c23}, \cite{c37}]. Collecting learning curves can be costly due to the training time required to train large models on each of the training subsets. We chose to train a small variant of GPT-2 that was expected to be large enough (i.e., sufficient parameters) to overfit all of the training set splits and yet small enough to train in a reasonable amount of time -- at most about 32 hours per training subset on a single GPU. We configure our GPT-2 model variant as follows: Vocabulary size 50257 sub-word tokens, maximum sequence length 512 tokens, depth 6 transformer blocks each with 8 self-attention heads and hidden dimension 512. The model has 44.9 million parameters total -- a rule of thumb in language modeling is to use a model with as many parameters as words in the largest dataset.

We train the models using the Adam optimizer with a static learning rate of 2e-4 and with batch sizes 12 and 24. The training objective is the cross-entropy loss of the model's prediction of the probability of the target next token in the input sentence. We empirically find that changing the batch size marginally changes the final loss ($<0.3\%$ change in cross-entropy), so we do not further explore optimization hyperparameters to mitigate total training time. Finally, we validate the models using the development dataset every 50-200 training steps, depending on the size of the dataset—smaller datasets require fewer training steps for the model to converge. We early-stop training when the development set loss stops improving for more than 15 validation runs.

\subsection{Evaluation Process and Effective Dataset Size}
Our objective is to measure how much the data distribution has changed over time. In that cause, we evaluable how well a dataset that has been sampled from one time period, can predict values for each other time period’s data. To do so, we train a model and evaluate its test error for each time period over multiple time periods. Furthermore, we characterize the learning curves so that we can translate measured test errors back to equivalent dataset sizes. Finally, we present the effectiveness curve.

In the training phase, we first find the finest model for each time period and each dataset size. The finest model is the one that achieves a smaller development set loss. Its selection process mimics the way models are chosen for deployment in AI-enabled products. To find the finest model, at each training run, we validate the models on the given time period’s development set and choose the model weights that achieve smaller development set loss. When we test with multiple different batch sizes, the finest model is the one that achieves superior performance in separate training runs for the given time period and training set size. 

We collect the finest model for each training set size ranging from 1.25 to 40 million words. Doing so allows us to construct learning curves across different time periods. We cross-evaluate all finest model -- one for each time period and training set size -- by evaluating them on the test sets for all other time periods. We use these results to curve fit learning curves and indirectly calculate its inverse: Given finest models for the time period $t_1$, and their evaluation scores for the time period, $t_0$ ($t_0$ can be equal to $t_1$), these scores will be used to show how increasing the training set size from period $t_1$ might improve prediction accuracy for the time period $t_0$. We curve fit learning curves with power-laws. 

Figure 4 shows examples of learning curves for models trained at different times. Each curve shows a model that has been trained on a specific time-period. The learning curves are different from each other and form parallel curves. The offset is due to change in the entropy $H(P)$, which is different at different times. Earlier models like those that have been trained in 2010 have lower values than the model of 2018. To answer why this is happening, we should look at Figure 3. As apparent from Figure 3, the dataset size per month is growing, which is a clear sign of the increase in the contribution and growth of the user base. This growth adds to the diversity in topics as well as language styles. The more diverse the dataset, the higher its entropy. It is also apparent from this graph that the learning curve is a decreasing function, and hence, more data causes lower cross-entropy value.

\begin{figure}[hbtp]
\centering
\includegraphics[scale=1]{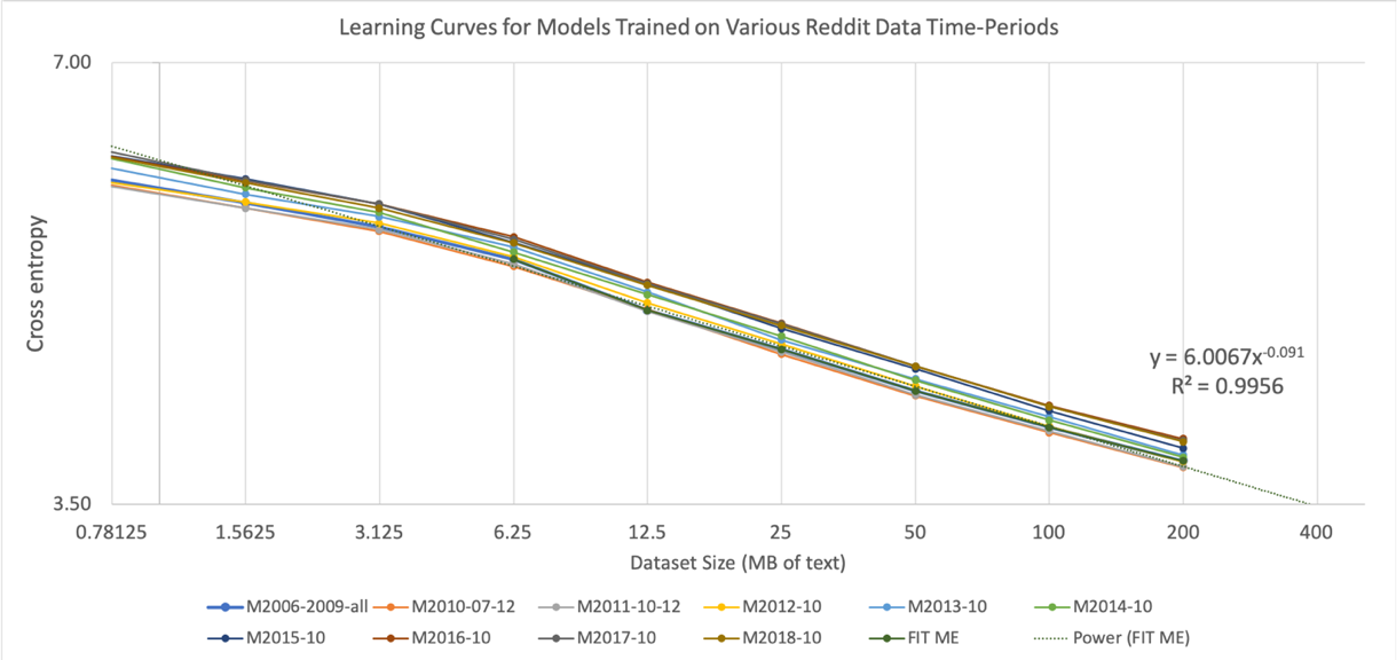}
\caption{Measured learning curves for models that have been trained at different times. The x-axis is in the log-scale and shows the dataset size. Y-axis is the cross-entropy value. The legend describes the time we used to train these models. For example, the yellow curve shows a model that has been trained on data from October 2012.}
\end{figure}

\begin{figure}[hbtp]
\centering
\includegraphics[scale=1]{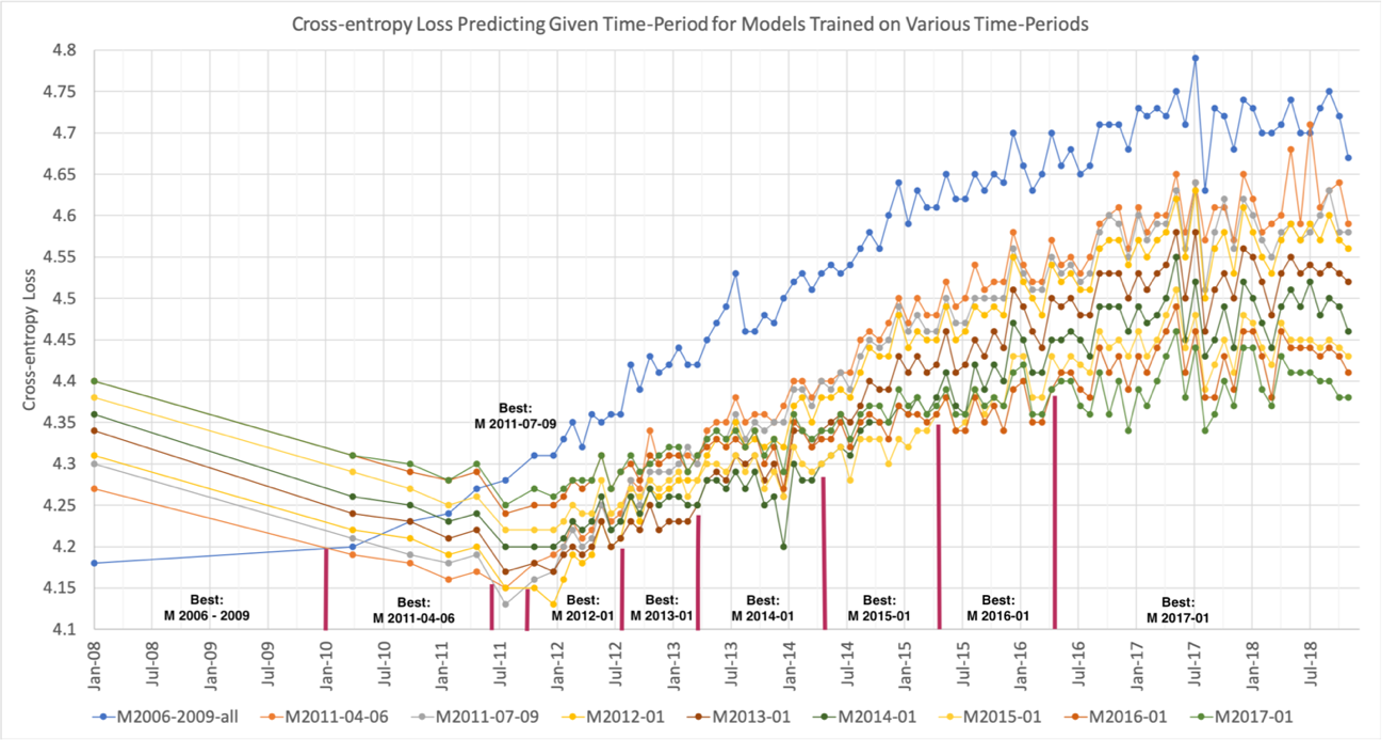}
\caption{Cross entropy loss value when we use a model that has been trained on year z (each curve) and is tested on data from year x (x-axis). Y-axis is the cross-entropy loss. The legend describes the time we used to train these models. For example, the green curve shows a model that has been trained on data from January 2014. The best cross-entropy loss in each time period is mentioned in this graph as well.}
\end{figure}

Figure 5 shows test evolution results for models trained on different time periods. Training size is fixed, and the algorithm is trained on data from a few time-periods. Periods are shown in the legend section of this Figure. Each point in this graph is the evaluation result of a training and test pair and curves are made by joining pairs with similar training time. For example, the blue curve shows the finest model's test results that have been trained 2006-2009 and tested on every other time. 

The first observation is that the best model for prediction in $t_0$ is the one trained on data from $t_0$. As an example, before January 2010, the model that has been trained on data from 2006-2009 ($m_{2006-2009}$) has the lowest cross-entropy and hence, has the best predicting power compared to other curves. In contrast, from January 2010 to June 2011, the April 2010’s model ($m_{2010-04-06}$) is the best performer replacing the blue curve. It immediately shows perishability. It is because the best performing model at one period loses its power as we move away from its sampling time. Despite apparent perishability, as time goes by, we see an increase in the cross-entropy values across all models. It is again due to the increase in the diversity of topics in Reddit data over time. In other words, the entropy function $H(P)$ is increasing.

Finally, we invert these learning curves to estimate the equivalent dataset size from time period $t_1$ when predicting for the time period $t_0$. Start with the best model, $m_{t_1,50M}$, for time period $t_1$ trained on 50 million words, for example. Evaluate $m_{t_1,50M}$ to collect cross-entropy loss for time period $t_0$. Now use the learning curve for models trained and tested on time period $t_0$ to estimate how much training data from time period $t_0$ is required to achieve that cross-entropy loss. Suppose the inverted learning curve yields 40 million words required in time period $t_0$, then the equivalent dataset size from time period $t_1$ is 40 million words at time $t_0$, or it is effectively 80\% of its time $t_1$ size. 

Figure 6 shows the equivalent dataset sizes for models trained on the 100MB of data sampled from different times. We chose 100MB for this graph to make it easier for readers to convert values to percentages. As seen in this Figure, for periods after sampling time, the equivalent sizes are monotonically decreasing. Despite overall monotonicity, we need to answer two questions about this graph:

\begin{enumerate}
\item Why do we observe higher equivalence variability on curves with higher equivalence (Closer to 100MB) sizes?
\item Why do we, on some occasions, observe a sudden increase in all equivalence curves?
\end{enumerate}

\begin{figure}[hbtp]
\centering
\includegraphics[scale=1]{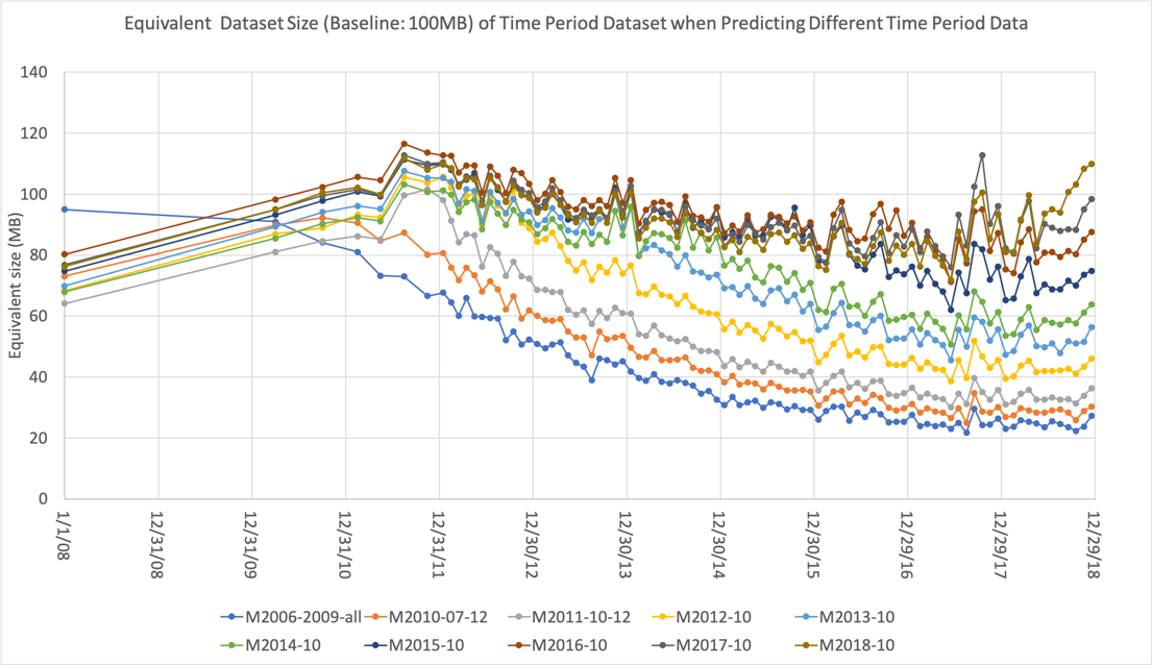}
\caption{Equivalent sizes over time (x-axis) when we used 100MB of data in the training phase. Each curve is the trained model. The legend describes the time we used to train these models. For example, the yellow curve shows a model that has been trained on data from October 2012.}
\end{figure}

For the first question, we believe it happens due to numerical errors in the inversion of learning curves. As we see in Figure 4, learning curves have power-law functional forms. Hence, in different regions of the learning curve, small change in measured cross-entropy translates to different magnitudes of change in equivalent sizes. For example, in Figure 4, if the training size is 100MB with measured cross-entropy of 5, the equivalent size is roughly 25MB. A small change of 0.1 in the measured cross-entropy translates to an equivalent size of roughly 20MB, which is 5MB different from the previous measurement. However, a similar small change, when the cross-entropy is 4, makes the difference of roughly 50MB. Therefore, the closer the equivalent size is to the training size, a smaller error causes a higher variability. This also explains the overshoots of later models (2017 and 2018) in equivalent sizes in August 2017.

For the second question, aside from the test set's sampling issues, model errors, and numerical error in fitting the learning curve's functional, we believe it is natural for events on those occasions to be slightly more predictable by all models. For example, for August 2017, if we look at the predictive power of $m_{2006-2009}$, we cannot find a considerable change, and sudden increase looks normal. However, due to the magnification of error and variability in later models (Models with sampling time closer to 2017), we see considerable changes in their equivalence values that sometimes lead to overshoots above 100MB. 

At last, Figure 7 shows the effectiveness curves. To deal with issues of the sudden increase in equivalent sizes, we made a slight alteration on the way we calculate the effectiveness curve. In this way, since theoretically ${\bar{n}}_{D_{n,0}}=n$, we calculate $E_{n,t}=\frac{{\bar{n}}_{D_{n,t}}}{n}=\frac{{\bar{n}}_{D_{n,t}}}{{\bar{n}}_{D_{n,0}}}$. In other words, instead of dividing the equivalent size of time t to 100MB, we divide it by the measured equivalent size of test time. It is as if we divide the measured value by the value of the best model predicting the test time. Doing this process over models from a few time periods creates Figure 7.

In this Figure, we can confirm a monotonic decrease of the effectiveness curve. It is interesting to see that the effectiveness curves of models from different times are all lined up. As this graph shows, roughly around 7 to 8 years, the value of data for the algorithm and the next word prediction task drops 50\%. Furthermore, we can see small periodic behavior in the measurements.  For example, looking at the values of days 365, 730, and 1095 and comparing them with the values of days 181, 550, and 915, we can see small ripples in the overall form of effectiveness functional. It suggests small periodicity in the data.

\begin{figure}[hbtp]
\centering
\includegraphics[scale=1]{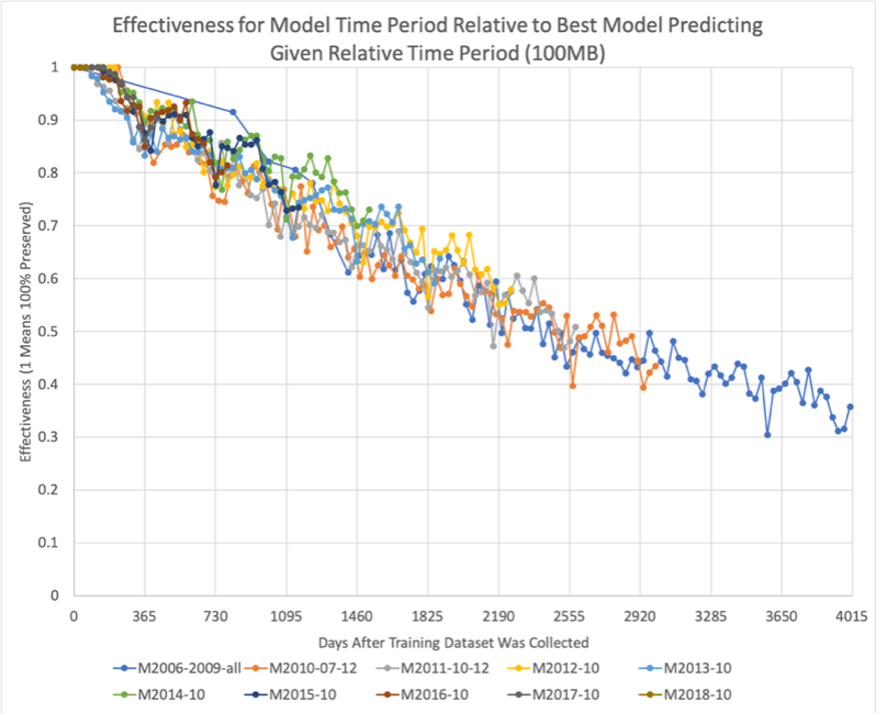}
\caption{Effectiveness curve. The X-axis shows the number of days after the training dataset was collected. The Y-axis shows the effectiveness of the trained model. 1 means that 100\% of the dataset's value has persevered. The legend describes the time we used to train these models. For example, the yellow curve shows a model that has been trained on data from October 2012. }
\end{figure}

\section{Conclusion}
An increase in the size of a dataset improves the generalizability and the accuracy of machine learning models. This argument and managerial theories on how the business output scales with the availability of resources convinced economists and data scientists that having more data always improves the quality of AI-based products and services.  For long, such statements triggered debates on whether stock of available data owned by big tech firms creates a barrier to the entry of competitors and if, much like the network effect, the data network effect creates a winner take all situation. Several empirical studies contested this view and argued that such economies of scale are hard to achieve in AI-based businesses for a variety of reasons. These empirical studies often cited data's diminishing return to scale and argued that the marginal value of new data points decreases as datasets size grows. 

 In our paper, we argued that time-dependency, despite having a significant effect on scaling the business value, is neglected in these debates. Time-dependency refers to the change in business value over time because of training models on datasets sampled from a dynamically changing environment. We cite the innovation in products and services space as well as the change in consumers' tastes and behavior as contributing factors to the change in environments over time. We further argue that due to innovation, with certainty, older datasets lose relevance over time. Mainly because the data-generating probability distributions in the future are different in that there is no combination of distributions from the past that can add up the future probability distributions. 

We theoretically show that even a perfect model trained on an infinite supply of time-dependent data may have lower accuracy than the same model trained on a recent (perfectly relevant) dataset of finite size. For dynamically changing business environments, this theorem immediately dismisses the role of stock of historical data in creating a barrier to entry of a competitor mainly because a competing firm equipped with a finite (yet sufficient) amount of recent data can attain a similar accuracy score and enter the market. In addition, this theorem has economic modeling implications. It states that a simple discounting function often used in the form of exponential decay over time is not suitable for modeling the growth in the data economy. Instead, an adequate discounting model should be a function of the size in addition to being a function of time.

We further introduced metrics like the \qq{equivalent size} that report the impact of time-dependency in dataset sizes. To evaluate the equivalent size, we first defined an oracle dataset. We then measured the size of the oracle dataset that, if used for training the machine learning model, leads to a similar accuracy score as the model trained on our dataset. In other words, the oracle dataset size creates a base of comparison in comparing various datasets. We then introduced the substitution function that measures the gain/loss in equivalent size when substituting datasets from various times. The equivalent size and the substitution function are the building blocks of our machinery to measure the value of data over time. Lastly, we proved the existence of \qq{equivalent time} to extend our machinery's capability to compare values of dataset curated over time. In our extended machinery, a dataset curated over time is represented by another dataset of similar size sampled from the equivalent time. Using the extended machinery, we formulated offloading algorithm. The outcome of this algorithm suggests that a business may remove old (less relevant) data from its repository and, despite losing size, end up in a better competitive position. In other words, the gain in relevance counterbalances the loss in size. On the other hand, a successful iteration of this algorithm suggests that increasing the dataset size by including an older dataset may put a firm in a disadvantageous position. 

The offloading algorithm builds the case for our argument that, in a rapidly changing environment, a firm should focus on the flow of data (defined as data collection rate) as the primary value driver instead of the stock of data. To formalize this idea, we first showed that when the flow of data increases, a firm gains/loses more if it substitutes data from various times. Then, we proved that the increase in the gain/loss (in the substitution function) makes the firm offload more often to bring the equivalent time closer to the prediction time. Such action makes the firm focus on the most relevant (recently collected) data. Therefore, when a firm grows (which often leads to an increase in the flow of data), it focuses on the most relevant (recent) data as the primary value driver. Consequently, we argue that the flow of data is the main value driver for big tech companies. Therefore, the marginal value of new data for such firms is always positive and economically significant, forcing them to collect new data aggressively.

Our findings can explain the mixed the result in the literature. First, we acknowledge the feedback loop logic in \cite{c24}. However, Proposition 1 shows that the stock of available data produced by the feedback loop has a finite oracle size because of time dependency. Hence, despite the accelerated growth in the size of data repository, we shouldn't expect a significant increase in created business value.  In other words, the feedback loop stalls in dynamically changing environments unless the firm offloads its less relevant data and focuses on the flow of data as the primary value driver. This finding supports the reported results in \cite{c13} and \cite{c5} since both search engine [\cite{c13}] and advertisement [\cite{c5}] businesses use time-sensitive data and hence, face significant time dependency. 

We can extend our arguments and results to study other data characteristics as long as we can model them by a change in the data-generating probability distribution. It is because all our definitions, theorems, and propositions are a function of variation in the distributions. For example, we may extend this result to measure the value loss in the user dimension. In other words, we may model the heterogeneity in preferences across users by variation in their preference distributions. Then, we can measure the value of a user's data on predicting other user's preferences. In this paper, we chose to center our arguments around the time dimension since it is easier to visualize value decline over time. Mainly because experimenting over time dimension has the benefit of having a possibly semi-monotonic decline in the value of data.

%
%
%



\bibliographystyle{informs2014} 
\bibliography{MS_Value_of_Data_V5.bib} 

\begin{thebibliography}{45}
\providecommand{\natexlab}[1]{#1}
\providecommand{\url}[1]{\texttt{#1}}
\providecommand{\urlprefix}{URL }

\bibitem[{Abrardi et~al.(2019)Abrardi, Cambini, \protect\BIBand{} Rondi}]{c1}
Abrardi L, Cambini C, Rondi L (2019) The economics of artificial intelligence:
  A survey. \emph{Robert Schuman Centre for Advanced Studies Research Paper No.
  RSCAS} 58.

\bibitem[{Aghion et~al.(2019)Aghion, Jones, \protect\BIBand{} Jones}]{c2}
Aghion P, Jones BF, Jones CI (2019) \emph{9. Artificial Intelligence and
  Economic Growth} (University of Chicago Press).

\bibitem[{Agrawal et~al.(2018)Agrawal, Gans, \protect\BIBand{} Goldfarb}]{c4}
Agrawal A, Gans J, Goldfarb A (2018) \emph{Prediction machines: the simple
  economics of artificial intelligence} (Harvard Business Press).

\bibitem[{Agrawal et~al.(2019)Agrawal, Gans, \protect\BIBand{} Goldfarb}]{c3}
Agrawal A, Gans J, Goldfarb A (2019) Economic policy for artificial
  intelligence. \emph{Innovation Policy and the Economy} 19(1):139--159.

\bibitem[{Arnold et~al.(2018)Arnold, Marcus, Petropoulos, \protect\BIBand{}
  Schneider}]{c5}
Arnold R, Marcus JS, Petropoulos G, Schneider A (2018) \emph{Is data the new
  oil? Diminishing returns to scale} (Calgary: International Telecommunications
  Society (ITS)).

\bibitem[{Bajari et~al.(2018)Bajari, Chernozhukov, Horta{\c{c}}su,
  \protect\BIBand{} Suzuki}]{c6}
Bajari P, Chernozhukov V, Horta{\c{c}}su A, Suzuki J (2018) The impact of big
  data on firm performance: An empirical investigation. Technical report,
  National Bureau of Economic Research.

\bibitem[{Baldwin(2019)}]{c7}
Baldwin R (2019) \emph{The globotics upheaval: Globalization, robotics, and the
  future of work} (Oxford University Press).

\bibitem[{Begenau et~al.(2018)Begenau, Farboodi, \protect\BIBand{}
  Veldkamp}]{c8}
Begenau J, Farboodi M, Veldkamp L (2018) Big data in finance and the growth of
  large firms. \emph{Journal of Monetary Economics} 97:71--87.

\bibitem[{Bergemann et~al.(2020)Bergemann, Bonatti, \protect\BIBand{} Gan}]{c9}
Bergemann D, Bonatti A, Gan T (2020) \emph{The economics of social data}
  (Cowles Foundation discussion paper).

\bibitem[{Brynjolfsson et~al.(2018)Brynjolfsson, Mitchell, \protect\BIBand{}
  Rock}]{c10}
Brynjolfsson E, Mitchell T, Rock D (2018) What can machines learn, and what
  does it mean for occupations and the economy? \emph{AEA Papers and
  Proceedings}, volume 108, 43--47.

\bibitem[{Carriere-Swallow \protect\BIBand{} Haksar(2019)}]{c11}
Carriere-Swallow MY, Haksar MV (2019) \emph{The economics and implications of
  data: an integrated perspective} (International Monetary Fund).

\bibitem[{Casella \protect\BIBand{} Berger(2021)}]{c12}
Casella G, Berger RL (2021) \emph{Statistical inference} (Cengage Learning).

\bibitem[{Chiou \protect\BIBand{} Tucker(2017)}]{c13}
Chiou L, Tucker C (2017) Search engines and data retention: Implications for
  privacy and antitrust. Technical report, National Bureau of Economic
  Research.

\bibitem[{Claussen et~al.(2021)Claussen, Peukert, \protect\BIBand{} Sen}]{c14}
Claussen J, Peukert C, Sen A (2021) The editor and the algorithm: Returns to
  data and externalities in online news. \emph{Available at SSRN 3479854} .

\bibitem[{Cockburn et~al.(2019)Cockburn, Henderson, \protect\BIBand{}
  Stern}]{c15}
Cockburn IM, Henderson R, Stern S (2019) \emph{4. The Impact of Artificial
  Intelligence on Innovation: An Exploratory Analysis} (University of Chicago
  Press).

\bibitem[{Cowgill \protect\BIBand{} Tucker(2020)}]{c16}
Cowgill B, Tucker CE (2020) Algorithmic fairness and economics. \emph{Columbia
  Business School Research Paper} .

\bibitem[{Cr{\'e}mer et~al.(2019)Cr{\'e}mer, de~Montjoye, \protect\BIBand{}
  Schweitzer}]{c17}
Cr{\'e}mer J, de~Montjoye YA, Schweitzer H (2019) Competition policy for the
  digital era. \emph{Report for the European Commission} .

\bibitem[{De~Corniere \protect\BIBand{} Taylor(2020)}]{c18}
De~Corniere A, Taylor G (2020) \emph{Data and competition: a general framework
  with applications to mergers, market structure, and privacy policy} (CEPR
  Discussion Paper No. DP14446).

\bibitem[{Fan et~al.(2019)Fan, Jernite, Perez, Grangier, Weston,
  \protect\BIBand{} Auli}]{c19}
Fan A, Jernite Y, Perez E, Grangier D, Weston J, Auli M (2019) Eli5: Long form
  question answering. \emph{arXiv preprint arXiv:1907.09190} .

\bibitem[{Farboodi et~al.(2019)Farboodi, Mihet, Philippon, \protect\BIBand{}
  Veldkamp}]{c20}
Farboodi M, Mihet R, Philippon T, Veldkamp L (2019) Big data and firm dynamics.
  \emph{AEA papers and proceedings}, volume 109, 38--42.

\bibitem[{Farboodi \protect\BIBand{} Veldkamp(2021)}]{c21}
Farboodi M, Veldkamp L (2021) A growth model of the data economy. Technical
  report, National Bureau of Economic Research.

\bibitem[{Furman et~al.(2019)Furman, Coyle, Fletcher, McAuley,
  \protect\BIBand{} Marsden}]{c22}
Furman J, Coyle D, Fletcher A, McAuley D, Marsden P (2019) Unlocking digital
  competition: Report of the digital competition expert panel. \emph{UK
  government publication, HM Treasury} .

\bibitem[{GPT-2(2018-2020)}]{c23}
GPT-2 (2018-2020) Gpt-2 source code: https://github.com/openai/gpt-2. OpenAI.

\bibitem[{Gregory et~al.(2021)Gregory, Henfridsson, Kaganer, \protect\BIBand{}
  Kyriakou}]{c24}
Gregory RW, Henfridsson O, Kaganer E, Kyriakou H (2021) Data network effects:
  Key conditions, shared data, and the data value duality. \emph{Academy of
  Management Review} .

\bibitem[{Hagiu \protect\BIBand{} Wright(2020)}]{c27}
Hagiu A, Wright J (2020) Data-enabled learning, network effects and competitive
  advantage. \emph{working paper} .

\bibitem[{Hestness et~al.(2017)Hestness, Narang, Ardalani, Diamos, Jun,
  Kianinejad, Patwary, Ali, Yang, \protect\BIBand{} Zhou}]{c25}
Hestness J, Narang S, Ardalani N, Diamos G, Jun H, Kianinejad H, Patwary M, Ali
  M, Yang Y, Zhou Y (2017) Deep learning scaling is predictable, empirically.
  \emph{arXiv preprint arXiv:1712.00409} .

\bibitem[{Holtz et~al.(2020)Holtz, Carterette, Chandar, Nazari, Cramer,
  \protect\BIBand{} Aral}]{c26}
Holtz D, Carterette B, Chandar P, Nazari Z, Cramer H, Aral S (2020) The
  engagement-diversity connection: Evidence from a field experiment on spotify.
  \emph{Proceedings of the 21st ACM Conference on Economics and Computation},
  75--76.

\bibitem[{Ichihashi(2021)}]{c28}
Ichihashi S (2021) The economics of data externalities. \emph{Journal of
  Economic Theory} 196:105316.

\bibitem[{Jones \protect\BIBand{} Tonetti(2020)}]{c29}
Jones CI, Tonetti C (2020) Nonrivalry and the economics of data. \emph{American
  Economic Review} 110(9):2819--58.

\bibitem[{Korinek \protect\BIBand{} Stiglitz(2019)}]{c30}
Korinek A, Stiglitz JE (2019) \emph{14. Artificial Intelligence and Its
  Implications for Income Distribution and Unemployment} (University of Chicago
  Press).

\bibitem[{Kullback \protect\BIBand{} Leibler(1951)}]{c31}
Kullback S, Leibler RA (1951) On information and sufficiency. \emph{The annals
  of mathematical statistics} 22(1):79--86.

\bibitem[{Lambrecht \protect\BIBand{} Tucker(2015)}]{c32}
Lambrecht A, Tucker CE (2015) Can big data protect a firm from competition?
  \emph{Available at SSRN 2705530} .

\bibitem[{Milgrom \protect\BIBand{} Tadelis(2019)}]{c33}
Milgrom PR, Tadelis S (2019) \emph{23. How Artificial Intelligence and Machine
  Learning Can Impact Market Design} (University of Chicago Press).

\bibitem[{Newman(2014)}]{c34}
Newman N (2014) Search, antitrust, and the economics of the control of user
  data. \emph{Yale J. on Reg.} 31:401.

\bibitem[{Petit(2017)}]{c35}
Petit N (2017) {Antitrust and Artificial Intelligence: A Research Agenda}.
  \emph{Journal of European Competition Law \& Practice} 8(6):361--362, ISSN
  2041-7764, \urlprefix\url{http://dx.doi.org/10.1093/jeclap/lpx033}.

\bibitem[{Prufer \protect\BIBand{} Schottmuller(2017)}]{c36}
Prufer J, Schottmuller C (2017) Competing with big data. \emph{TILEC Discussion
  Paper} .

\bibitem[{Radford et~al.(2019)Radford, Wu, Child, Luan, Amodei, Sutskever
  et~al.}]{c37}
Radford A, Wu J, Child R, Luan D, Amodei D, Sutskever I, et~al. (2019) Language
  models are unsupervised multitask learners. \emph{OpenAI blog} 1(8):9.

\bibitem[{Reimers \protect\BIBand{} Shiller(2018)}]{c38}
Reimers I, Shiller B (2018) Welfare implications of proprietary data
  collection: an application to telematics in auto insurance. \emph{Available
  at SSRN 3125049} .

\bibitem[{Rubinfeld \protect\BIBand{} Gal(2017)}]{c39}
Rubinfeld DL, Gal MS (2017) Access barriers to big data. \emph{Ariz. L. Rev.}
  59:339.

\bibitem[{Schaefer et~al.(2018)Schaefer, Sapi, \protect\BIBand{} Lorincz}]{c40}
Schaefer M, Sapi G, Lorincz S (2018) The effect of big data on recommendation
  quality: The example of internet search. \emph{DIW Berlin Discussion Paper} .

\bibitem[{Shannon(1948)}]{c41}
Shannon CE (1948) A mathematical theory of communication. \emph{The Bell system
  technical journal} 27(3):379--423.

\bibitem[{Tirole(2020)}]{c42}
Tirole J (2020) Competition and the industrial challenge for the digital age.
  \emph{paper for IFS Deaton Review on Inequalities in the Twenty-First
  Century} .

\bibitem[{Van~Til et~al.(2017)Van~Til, Van~Gorp, \protect\BIBand{} Price}]{c43}
Van~Til H, Van~Gorp N, Price K (2017) Big data and competition. \emph{Ecorys
  Study for the Dutch Ministry of Economic Affairs, Ecorys, Rotterdam.
  https://www. rijksoverheid.
  nl/binaries/rijksoverheid/documenten/rapporten/2017/06/13/big-data-and-competition/big-data-andcompetition.
  pdf} .

\bibitem[{Varian(2019)}]{c44}
Varian H (2019) \emph{16. Artificial Intelligence, Economics, and Industrial
  Organization} (University of Chicago Press).

\bibitem[{Wang et~al.(2018)Wang, Singh, Michael, Hill, Levy, \protect\BIBand{}
  Bowman}]{c45}
Wang A, Singh A, Michael J, Hill F, Levy O, Bowman SR (2018) Glue: A multi-task
  benchmark and analysis platform for natural language understanding.
  \emph{arXiv preprint arXiv:1804.07461} .

\end{thebibliography}

\newpage
\section*{APPENDIX}
\subsection*{Proof of Theorem 1)}
Define $v=-\log(m\left(d,\theta\right))$. For a given $\theta$ and IID $d_i\sim P(\omega)$, $v_i$ becomes IID samples of random variable $v$. If $Ev_i^2<\infty$, for a large number of data points we can use central limit theorem and hence, 
\[\frac{1}{n}\sum_{i=1}^{n}v_i=E_P\left(v\right)+o\left(\frac{C_1}{\sqrt n}\right)\mathcal{N}(0,1)\]
Where $C_1$ is a function of $var(v)$. Note that \[E_P\left(v\right)=\ -E_P\left(\log{\left(m\left(d,\theta\right)\right)}\right)=-E_P\left(\log{\left(P\right)}\right)+E_P\left(\log{\left(P\right)}\right)-E_P\left(\log{\left(m\left(d,\theta\right)\right)}\right)=\]\[\ -E_P\left(\log{\left(P\left(d\right)\right)}\right)+E_P\log{\left(\frac{P(d)}{m(d,\theta)\ }\right)=H\left(P\right)+KL(P\ |\left|m\left(d,\theta\right)\right)}.\]
Therefore, 
\[-\frac{1}{n}\sum_{i=1}^{n}\log{\left(m\left(d_i,\theta\right)\right)}=H\left(P\right)+KL(\ P|\left|m\left(d,\theta\right)\ \right)+O\left(\frac{C_1}{\sqrt n}\right)\mathcal{N}\left(0,1\right)\ \]

Q.E.D.

\subsection*{Proof of Proposition 1)}
From our assumptions in the paper and the asymptotic efficiency of MLE [\cite{c12}], we know that $\lim_{n\rightarrow\infty}{m\left(d,\theta_n\right)=P(d)}$ where $\theta_n=\argmax_\theta{\sum_{i=1}^{n}\log{\left(m\left(d_i,\theta\right)\right)}}$.
Hence, for $E|\log{\left(m\left(d_i,\theta_n\right)\right)|<\infty\ }$ and using the strong law of large number we have
\[\lim_{n\rightarrow\infty}-\frac{1}{n}\sum_{i=1}^{n}{\log{\left(m\left(d_i,\theta_n\right)\right)}=H\left(P\right)+KL(P||m\left(d,\theta_\infty\right)){=H\left(P\right)+KL(P||P)=H(P)}}\]
Therefore, a model that has been trained on $D_{\infty,0}$ should reach the loss value $H\left(P_0\right)$. Assume $d^{(0)}\sim P_0\left(d\right)$ and $d^{(t)}\sim P_t\left(d\right)$. Consider a model that has been trained on a dataset from time $t$ ($D_{\infty,t}$) and been tested on a dataset from time 0,  $D_{\infty,0}$.  In this case, $\lim_{n\rightarrow\infty}{m\left(d^{\left(t\right)},\theta_n\right)=P_t\left(d\right)}$ where $\theta_{n,t}=\argmax_\theta{\sum_{i=1}^{n}\log{\left(m\left(d_i^{(t)},\theta\right)\right)}}$

The test loss value for this model is
\[\lim_{n\rightarrow\infty}-\frac{1}{n}\sum_{i=1}^{n}{\log{\left(m\left(d_i^{\left(0\right)},\theta_{\infty,t}\right)\right)}=H\left(P\left(d\right)\right)+KL\left(P(d)||\ m\left(d,\theta_{\infty,t}\right)\right)=H\left(P_0\right)}+KL(P_0||P_t)  \]
Since both $H\left(P_0\right)$ and $KL(P_0||P_t)$ are non-negative functions of distributions [\cite{c31},\cite{c41}], we conclude that the loss value is higher than $H\left(P_0\right)$. Therefore, a bounded size dataset should reach the loss value $H\left(P_0\right)+KL(P_0||P_t)$. 

Formalizing this argument, we define a neighborhood around $H(P_0)$ with the size $\delta>0$ and prove that with probability $(1-\epsilon)$,  any dataset of bounded size reaches a value in the neighborhood.

Mathematically, for large dataset samples $n\gg1$ and $\delta>0$, using theorem 1 we have
\[P\left(\left|-\frac{1}{n}\sum_{i=1}^{n}\log{\left(m\left(d_i^{\left(0\right)},\theta_{n,0}\right)\right)}-H\left(P_0\right)\right|>\delta\right)=\]\[P\left(\left|KL\left(P_0{||P}_t\right)+O\left(\frac{1}{\sqrt n}\right)\mathcal{N}\left(0,1\right)\right|>\delta\right)=P\left(\left|\mathcal{N}\left(KL\left(P_0{||P}_t\right),o\left(\frac{1}{\sqrt n}\right)\right)\right|>\delta\right)=\]\[
=P\left(\mathcal{N}\left(KL\left(P_0{||P}_t\right)-\delta,o\left(\frac{1}{\sqrt n}\right)\right)>0\right)+P\left(\mathcal{N}\left(KL\left(P_0{||P}_t\right)+\delta,o\left(\frac{1}{\sqrt n}\right)\right)<0\right)\]\[={\underbrace{\Phi\left(\frac{\delta-KL(P_0||P_t)}{o\left(\frac{1}{\sqrt n}\right)}\right)}}_{(i)}+{\underbrace{\Phi\left(\frac{-\delta-KL(P_0||P_t)}{o\left(\frac{1}{\sqrt n}\right)}\right)}}_{(ii)}\]
Where $\Phi(.)$ is the cumulative distribution function of standard Normal. In above equation, since $\delta>0$, (i) is bigger than (ii) which means
\[P\left(\left|-\frac{1}{n}\sum_{i=1}^{n}\log{\left(m\left(d_i^{\left(0\right)},\theta_{n,0}\right)\right)}-H(P_0)\right|>\delta\right)<2\Phi\left(\frac{\delta-KL(P_0||P_t)}{o\left(\frac{1}{\sqrt n}\right)}\right)\]

Since for $\delta<D(P_0{||P}_t)$ the numerator is negative, 
\[\lim_{n\rightarrow\infty}{\Phi\left(\frac{\delta-KL(P_0||P_t)}{o\left(\frac{1}{\sqrt n}\right)}\right)=\Phi\left(-\infty\right)=0}\]
Therefore, For any $\epsilon,\delta>0,\ \exists\ n_0<\infty\ \ s.t.\ \forall\ n>n_0\ $
\[P\left(\left|-\frac{1}{n}\sum_{i=1}^{n}\log{\left(m\left(d_i^{\left(0\right)},\theta_{n,0}\right)\right)}-H(P_0)\right|>\delta\right)<2\Phi\left(\frac{\delta-KL(P_0||P_t)}{o\left(\frac{1}{\sqrt n}\right)}\right)<\epsilon\]
Meaning that a dataset size of $n>n_0$ with probability $1-\epsilon$ surpass the performance of infinite dataset size from time $t$.

Q.E.D.

\subsection*{Proof of Proposition 2)}
As explained in the main text, the substitution function $f_n(t_1,t_2)$ measures the gain in substituting a dataset of size n from time $t_2$ with a dataset of same size from time $t_1$. We also proved in theorem 2 (a) that the substitution function is non-negative. In proving the claims, we have two additional assumptions. First, we assume that the monotonicity result proved in theorem 2 (b) is valid for all dataset sizes meaning that $f_n(t_1,t_2)$ is monotonic for all n. Second, we assume that $f_1\left(t_1,t_2\right)=1$ for all $t_1,t_2\in\mathbb{R}^+\cup{0}$. It is intuitive since, in our model, all elements have non-zero probability and hence, one data point carries in expectation same amount of information regardless of when it was sampled.

Now, fixing $t_2$, for all ${t\in\mathbb{R}}^+\cup{0}$ and all $n\in\mathbb{N}$ we have either

\begin{itemize}

\item $f_n\left(t,t_2\right)<f_\infty(t,t_2)$ which due to monotonicity means that it is increasing, and the claim is proved.
	
\item $f_n\left(t,t_2\right)>f_\infty(t,t_2)$ which means that for $t=0$ and $n<\infty, n>f_n\left(0,t_2\right)>f_\infty\left(0,t_2\right)\Longrightarrow1>f_\infty\left(0,t_2\right)=\infty$ which is a contradiction.
	
\item Or $f_n\left(t,t_2\right)$ intersects with $f_\infty\left(t,t_2\right)$ in a few points. Suppose $t_0$ is an intersection point. Since $f_n(t_0,t_2)$ is monotone in $n$ and $f_n\left(t_0,t_2\right)=f_\infty\left(t_0,t_2\right)=c$, for all $n$, $f_n\left(t_0,t_2\right)=c$.  If $c\neq1$, then $f_1\left(t_0,t_2\right)\neq1$ which is a contradiction and hence $c=1$. Between the intersection points if $ f_\infty\left(t,t_2\right)>1$, $f_n\left(t,t_2\right)$ is increasing since it is monotonic between $f_1\left(t,t_2\right)=1$ and $f_\infty\left(t,t_2\right)$ and if $f_\infty\left(t,t_2\right)<1, f_n\left(t,t_2\right)$ is decreasing since it is monotonic between $f_1\left(t,t_2\right)=1$ and $f_\infty\left(t,t_2\right)$.

\end{itemize}

In conclusion, $f_n(t_1,t_2)$ is increasing in $n$ if $f_n\left(t_1,t_2\right)>1$ and it is decreasing in $n$ if $f_n\left(t_1,t_2\right)<1$.  Hence, the substitution gain/loss increases with $n$. Because of Proposition 3 which states that a dataset curates over time has a loss value equal to a dataset of same size $\left(n\right)$ that is sampled from the equivalent time, we argue that $n$ is equivalent to the flow of data. Hence, the substitution gain/loss increases with the flow of data.

Q.E.D.

\subsection*{Proof of Theorem 2)}
\begin{itemize}
\item[a, c)] This is a direct result of theorem 1 and proposition1.
\item [b)] Due to monotonic decline of effectiveness over time, $KL\left(P_0||P_{t_1}\right)<KL(P_0||P_{t_2})$ for $t_2>t_1$ and $KL\left(P_0||P_{t_1}\right)>KL(P_0||P_{t_2})$ for $t_2<t_1$.

For sufficiently large number of datapoints, the model $m(d,\theta)$ almost converged to $P(d)$. Therefore, due to continuity and differentiability of the learning curve, we can use Taylor expansion of learning curve's inverse in the neighborhood of $P(d)$.
\[{\bar{n}}_{D_{n,t}}=EG_0^{-1}\left(H\left(P_0\right)+KL\left(P_0||m\left(d,\theta_{n,t}\right)\right)\ \right)\sim\]\[ G_0^{-1}\left(H\left(P_0\right)+KL\left(P_0{||P}_t\right)\right)\]\[+E\left[\left(KL\left(P_0||m\left(d,\theta_{n,t}\right)\right)-KL\left(P_0{||P}_t\right)\right)\frac{\partial G_0^{-1}\left(q\right)}{\partial q}|_{H\left(P_0\right)+KL\left(P_0{||P}_t\right)}\right]\ \]\[=\ G_0^{-1}\left(H\left(P_0\right)+KL\left(P_0{||P}_t\right)\right)-E\left[\left(E_{P_0}\log{\frac{m\left(d,\theta_{n,t}\right)}{P_t\left(d\right)}}\right)\right]\frac{\partial G_0^{-1}\left(q\right)}{\partial q}|_{H\left(P_0\right)+KL\left(P_0{||P}_t\right)}\]
Using Taylor expansion $\log\left(1+x\right) \sim x-\frac{x^2}{2}+\frac{x^3}{3}+o(x^4)$, in the neighborhood of $x=0$. We do this because we expect $m\left(d,\theta_{n,t}\right)\rightarrow P_t(d)$. Using Taylor expansion, we have 
\[n_{D_{n,t}}=G_0^{-1}\left(H\left(P_0\right)+KL\left(P_0{||P}_t\right)\right)\]\[-E_{P_0}\left(\frac{m\left(d,\theta_{n,t}\right)-P_t(d)}{P_t\left(d\right)}-\frac{1}{2}\left(\frac{m\left(d,\theta_{n,t}\right)-P_t\left(d\right)}{P_t\left(d\right)}\right)^2+\frac{1}{3}\left(\frac{m\left(d,\theta_{n,t}\right)-P_t\left(d\right)}{P_t\left(d\right)}\right)^3+o\left(\frac{m\left(d,\theta_{n,t}\right)-P_t(d)}{P_t\left(d\right)}\right)^4\right)\]\[\times \frac{\partial G_0^{-1}\left(q\right)}{\partial q}|_{H\left(P_0\right)+KL\left(P_0{||P}_t\right)}\]
Assuming $m(d,\theta)$ to be a continuous function of $\theta$, we can use theorem 10.1.12 in \cite{c12} (Asymptotic efficiency of MLE) and approximate $m\left(d,\theta_{n,t}\right)$ with respect to randomization in algorithms and choice of dataset in the training phase. Therefore,
\[m\left(d,\theta_{n,t}\right)\sim P_t(d)+\frac{1}{\sqrt n}\mathcal{N}(0,v\left(\theta\right))\]
Where $v(\theta)$ is the Cramer-Rao lower bound.
\[{\bar{n}}_{D_{n,t}}=G_0^{-1}\left(H\left(P_0\right)+KL\left(P_0{||P}_t\right)\right)\]\[-E\Bigg[E_{P_0}\left(\frac{1}{\sqrt n}\mathcal{N}\left(0,\frac{v\left(\theta\right)}{P_t\left(d\right)}\right)-\frac{1}{2n}\left(\mathcal{N}\left(0,\frac{v\left(\theta\right)}{P_t\left(d\right)}\right)\right)^2+\frac{1}{3n\sqrt n}\left(\mathcal{N}\left(0,\frac{v\left(\theta\right)}{P_t\left(d\right)}\right)\right)^3+o\left(\frac{1}{n^2}\right)\right)\Bigg] \]\[\times \frac{\partial G_0^{-1}\left(q\right)}{\partial q}|_{H\left(P_0\right)+KL\left(P_0{||P}_t\right)}
\]\[=G_0^{-1}\left(H\left(P_0\right)+KL\left(P_0{||P}_t\right)\right)+\frac{1}{2n}\left[E{E_{P_0}\left(\mathcal{N}\left(0,\frac{v\left(\theta\right)}{P_t\left(d\right)}\right)\right)}^2+o\left(\frac{1}{n^2}\right)\right]\times \ \frac{\partial G_0^{-1}\left(q\right)}{\partial q}|_{H\left(P_0\right)+KL\left(P_0{||P}_t\right)}\ \]
Since the first and third moment of centered Gaussian distribution is equal to 0. 

\textit{As a side note, the argument inside the brackets is positive. Since \[\frac{\partial G_0^{-1}\left(q\right)}{\partial q}<0\] we conclude 
\[\frac{1}{2n}\left[E{E_{P_0}\left(\mathcal{N}\left(0,\frac{v\left(\theta\right)}{P_t\left(d\right)}\right)\right)}^2\right]\ \frac{\partial G_0^{-1}\left(q\right)}{\partial q}|_{H\left(P_0\right)+KL\left(P_0{||P}_t\right)}<0\]
 Hence, ${\bar{n}}_{D_{n,t}}$  is an increasing function in n for sufficiently large $n$.}
 
Back to the prove, we now take the derivative of $f_n\left(t_1,t_2\right)$ with respect to $n$. For large $n$ we use the following approximation
\[{\hat{f}}_n\left(t_1,t_2\right)=\frac{G_0^{-1}\left(H\left(P_0\right)+KL\left(P_0{||P}_{t_1}\right)\right)+\frac{1}{2n}\left[E{E_{P_0}\left(\mathcal{N}\left(0,\frac{v\left(\theta\right)}{P_{t_1}\left(d\right)}\right)\right)}^2\right]\ \frac{\partial G_0^{-1}\left(q\right)}{\partial q}|_{H\left(P_0\right)+KL\left(P_0{||P}_{t_1}\right)}}{G_0^{-1}\left(H\left(P_0\right)+KL\left(P_0{||P}_{t_2}\right)\right)+\frac{1}{2n}\left[E{E_{P_0}\left(\mathcal{N}\left(0,\frac{v\left(\theta\right)}{P_{t_2}\left(d\right)}\right)\right)}^2\right]\ \frac{\partial G_0^{-1}\left(q\right)}{\partial q}|_{H\left(P_0\right)+KL\left(P_0{||P}_{t_2}\right)}}\]\[=\frac{{\bar{n}}_{D_\infty,t_1}+\frac{1}{2n}\left[E{E_{P_0}\left(\mathcal{N}\left(0,\frac{v\left(\theta\right)}{P_{t_1}\left(d\right)}\right)\right)}^2\right]\ \frac{\partial G_0^{-1}\left(q\right)}{\partial q}|_{H\left(P_0\right)+KL\left(P_0{||P}_{t_1}\right)}}{{\bar{n}}_{D_\infty,t_2}+\frac{1}{2n}\left[E{E_{P_0}\left(\mathcal{N}\left(0,\frac{v\left(\theta\right)}{P_{t_2}\left(d\right)}\right)\right)}^2\right]\ \frac{\partial G_0^{-1}\left(q\right)}{\partial q}|_{H\left(P_0\right)+KL\left(P_0{||P}_{t_2}\right)}}\]
To show the derivative sign, we focus on the for large n (Omitting o$\left(\frac{1}{n^3}\right)$)
\[\Longrightarrow num\left(\frac{\partial{\hat{f}}_n\left(t_1,t_2\right)}{\partial n}\right) \sim \frac{1}{2n^2}\Bigg({\bar{n}}_{D_\infty,t_1}\left[E{E_{P_0}\left(\mathcal{N}\left(0,\frac{v\left(\theta\right)}{P_{t_2}\left(d\right)}\right)\right)}^2\right]\ \frac{\partial G_0^{-1}\left(q\right)}{\partial q}|_{H\left(P_0\right)+D\left(P_0{||P}_{t_2}\right)}\]\[-{\bar{n}}_{D_\infty,t_2}\left[E{E_{P_0}\left(\mathcal{N}\left(0,\frac{v\left(\theta\right)}{P_{t_1}\left(d\right)}\right)\right)}^2\right]\ \frac{\partial G_0^{-1}\left(q\right)}{\partial q}|_{H\left(P_0\right)+D\left(P_0{||P}_{t_1}\right)}\Bigg)\]
Since the argument in the brackets are not a function of $n$, we can conclude that for large $n$, the substitution function $f_n\left(t_1,t_2\right)$ is monotonic in $n$. 
\end{itemize}
Q.E.D.

\subsection*{Proof of Lemma 1)}
Assume dataset $D_{n,t}$ is sampled over time with the density function $\lambda_{t=t_0}=\frac{1}{n}\sum_{i=1}^{n}{1(t_i=t_0)}$. Considering each sample a random variable, number of times event \qq{a} happens i.e. $1\left(d\in a\right)=1$ in the dataset is equal to $\sum_{i=1}^{n}{1_{t_i}(d\in a)}$. Therefore, the expected frequency of the event $\{d\in a\}$ is equal to
\[P_{D_{n,t}}\left(d\in a\right)=E\left(\sum_{i=1}^{n}\frac{1_{t_i}\left(d\in a\right)}{n}\right)\underbrace{=}_{Fubini\ theorem}{\frac{1}{n}\sum_{i=1}^{n}{{E(1}_{t_i}(d\in a))}}=\frac{1}{n}\sum_{i=1}^{n}{P_{t_i}(d\in a)}\]
Integrating the density function $\lambda_t$ into formulation
\[P_{n,\left[0,t\right],\lambda_t}\left(d\in a\right)=\frac{1}{n}\sum_{i=1}^{n}{P_{t_i}(d\in a)}=\frac{1}{n}\int_{0}^{t}\sum_{i=1}^{n}{P_s\left(d\in a\right)1\left(t_i=s\right)}\ ds\]\[=\int_{0}^{t}{P_s\left(d\in a\right)\frac{1}{n}\sum_{i=1}^{n}1\left(t_i=s\right)}\ ds=\int_{0}^{t}{P_s\left(d\in a\right)\lambda_sds}\]
Q.E.D.

\subsection*{Proof of Proposition 3)}
Using lemma 1, we know that dataset’s net distribution is
\[P_{\left[0,t\right],\lambda_t}\left(d\in a\right)=\int_{0}^{t}{P_s\left(d\in a\right)\lambda_sds}\]
Therefore, training on the dataset of infinite size and test it at time 0, the error will be equal to 
\[H\left(P_0\right)+KL\left(P_0{||P}_{\left[0,t\right],\lambda_t}\right)=H\left(P_0\right)+KL\left(P_0||\int_{0}^{t}{P_s\left(d\in a\right)\lambda_sds}\right)\]
Since KL-divergence is a convex function [31], we use Jensen inequality to derive an upper bound
\[KL\left(P_0||\int_{0}^{t}{P_s\left(d\in a\right)\lambda_sds}\right)=KL\left(\int_{0}^{t}{P_0\lambda_sds}||\int_{0}^{t}{P_s\left(d\in a\right)\lambda_sds}\right)\]\[=\int_0^t \lambda_s KL(P_0||P_s)ds \leq \max_{s\in[0,t]}KL(P_0||P_s)\]
Besides, we know that KL-divergence is non negative which means
\[KL\left(P_0||P_0\right)=0\le KL\left(P_0||\int_{0}^{t}{P_s\left(d\in a\right)\lambda_sds}\right)\le \max_{s\in[0,t]} KL(P_0||P_s)\]
Since we assumed in this paper that the function $h\left(t\right)=D\left(P_0{||P}_t\right)$ is continuous over time (The change in distribution is gradual and hence, $h(t)$ is continuous) There exist a time $t^\ast\in[0,t]$ such that 
\[KL\left(P_0{||P}_{t^\ast}\right)=KL\left(P_0||\int_{0}^{t}{P_s\left(d\in a\right)\lambda_sds}\right)\]
Therefore
\[H\left(P_0\right)+KL\left(P_0{||P}_{t^\ast}\right)=H\left(P_0\right)+KL\left(P_0||\int_{0}^{t}{P_s\left(d\in a\right)\lambda_sds}\right)\]
This means that $P_{t^\ast}$ generate the same loss value as $P_{\left[0,t\right],\lambda_t}$.

Q.E.D.

\subsection*{Proof of Proposition 4)}
Without loss of generality we focus our attention to the case of monotonic decline in the value of data. Increase in the substitution gain or loss means that the substitution gain becomes sharper as the dataset size increase due to an increase in the flow of data i.e. for all $t>t_2>t_1>0$, $\alpha\geq 1$, and for the flow rates $\psi_H (t) =\alpha \psi_L (t) >0$ 
\[ f_{n_H}\left(t_1,t_2\right)>f_{n_L}\left(t_1,t_2\right)\ \ \ \]
where
\[\begin{matrix}
n_H = \int_0^t \psi_H(t) dt \\
n_L = \int_0^t \psi_L (t) dt  \end{matrix}\]\[
\lambda_H (t) = \frac{\psi_H (t)}{n_H} = \frac{\alpha\psi_L(t)}{\alpha n_L} =\frac{\psi_L(t)}{ n_L} = \lambda_L(t)\\
\]

Since $\lambda_L(t)=\lambda_H(t)$, according to Lemma 1, the net distribution for datasets created from both flow rates $\psi_H$ and $\psi_L$ are identical. Therefore, according to Proposition 3, both distributions have identical equivalent times. Lets call the equivalent time for these datasets $t_2>0$. 

Remember
 the condition for a successful off-loading iteration from the equivalent time $t^\ast$ to $t^{\ast\ast}$ is \[ 
f_{n-n_0}\left(t^{\ast\ast},t^\ast\right)>\frac{n}{n-n_0}\]


In that case for all $t_1,t_2>0$ such that offloading for $n_L$ is feasible, i.e.
\[f_{n_L}\left(t_1,t_2\right)\geq\frac{n_L}{n_L-\int_{t_2}^t \psi_L(t)dt} \]
we have
\[f_{n_H}\left(t_1,t_2\right)>f_{n_L}\left(t_1,t_2\right)\geq\frac{n_L}{n_L-\int_{t_2}^t \psi_L(t)dt} =\frac{\alpha}{\alpha}\times \frac{n_L}{n_L-\int_{t_2}^t \psi_L(t)dt}=\frac{n_H}{n_H-\int_{t_2}^t \psi_H(t)dt} \]
\[\Rightarrow f_{n_H}\left(t_1,t_2\right)> \frac{n_H}{n_H-\int_{t_2}^t \psi_H(t)dt}\]
meaning that for all $t_1,t_2>0$ such that offloading is possible for the low flow rate $\psi_L(t)$, such offloading is also possible for high flow rate $\psi_H (t)$ and hence, the equivalent time for the high flow rate is weakly closer to the prediction time $0$ compared to the equivalnet time for low flow rate.
And that completes the proof.

Q.E.D.
 \newpage
\section{Appendix B}
We ran four experiments with different dataset sizes over Reddit data. The up-left Figure shows the effectiveness curve when we trained the model over 25MB of data. Up-right, down-left, and down-right show the curves for 50, 100, 200 MBs, respectively. As can be seen in these graphs, the effectiveness curve is becoming steeper as expected. Meaning that substitution gain will be monotonically increasing in the number of samples.
For example, looking at the effectiveness value for day 2920, we can see the effectiveness values of roughly 0.55, 0.5, 0.45, and 0.4 in the 25, 50, 100, and 200 MBs graphs, respectively. 
\[\begin{matrix}f_{25MB}\left(0,2920\right)\sim 1.81\ \\f_{50MB}\left(0,2920\right)\sim 2.00\\\begin{matrix}f_{100MB}\left(0,2920\right)\sim 2.22\\f_{200MB}\left(0,2920\right)\sim 2.50\\\end{matrix}\\\end{matrix}\]

\begin{figure}[hbtp]
\centering
\includegraphics[scale=1]{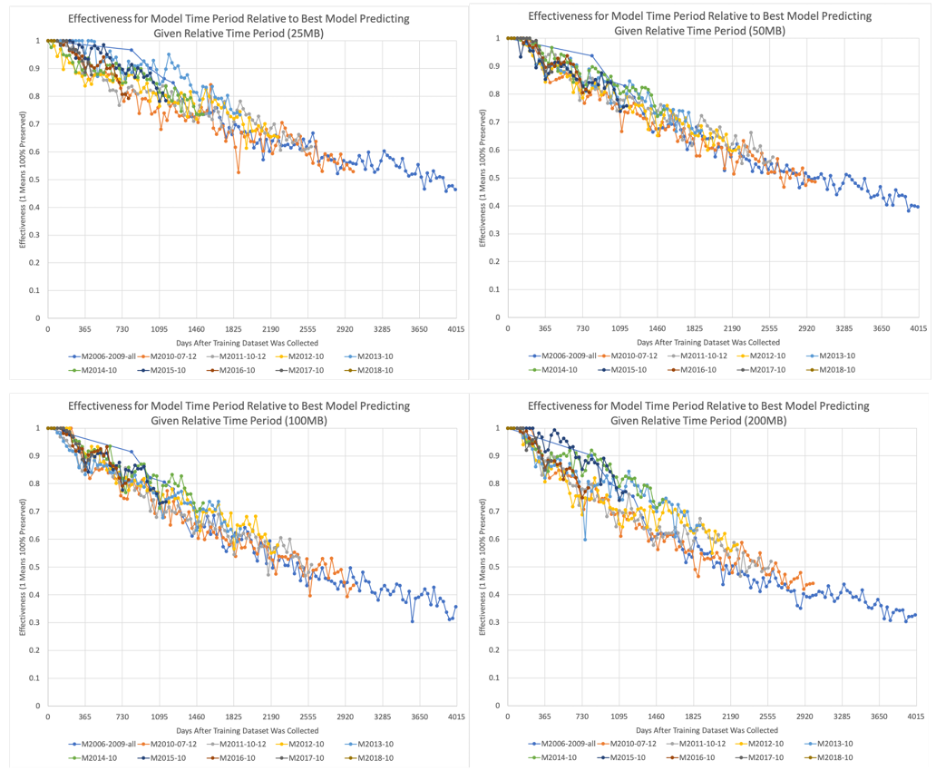}
\caption{Effectiveness graphs for various training sizes}
\end{figure}

\end{document}